\documentclass{article} % For LaTeX2e
\usepackage{iclr2027_conference,times}

\usepackage[utf8]{inputenc}
\usepackage[T1]{fontenc}
\usepackage{hyperref}
\usepackage{url}
\usepackage{booktabs}
\usepackage{amsfonts}
\usepackage{amsmath}
\usepackage{nicefrac}
\usepackage{microtype}
\usepackage{xcolor}
\usepackage{graphicx}
\usepackage{multirow}
\usepackage{subcaption}

\title{ProtSent: Protein Sentence Transformers}

\author{Dan Ofer$^{1,*}$ \quad Oriel Perets$^{2,*}$ \quad Michal Linial$^{1}$ \quad Nadav Rappoport$^{2}$ \\
$^{1}$Department of Biological Chemistry, The Hebrew University of Jerusalem \\
$^{2}$Stein Faculty of Computer and Information Science, Ben-Gurion University of the Negev \\
$^{*}$Equal contribution. \quad Correspondence: \texttt{orielpe@post.bgu.ac.il}
}

\iclrfinalcopy % Shows the author block (otherwise ICLR prints "Anonymous authors").
\begin{document}

\maketitle
% The ICLR style sets the running head inside \maketitle ("Published as a
% conference paper at ICLR 2027" in final mode); override it for the preprint.
\lhead{Preprint}

\begin{abstract}
%Intro
Protein language models produce rich per-residue representations, but the sequence embeddings obtained by mean-pooling them are never trained to place related proteins near one another, so the geometry that retrieval, clustering and nearest-neighbor transfer depend on is never optimized directly.
%Methods
We present ProtSent, a contrastive post-training framework that supervises a protein language model with three types of biological relation at once, Pfam family co-membership, AlphaFold~DB structural clusters and STRING interactions, and with deep mutational scanning fitness ordering as an optional fourth. It needs no structure at training or inference, trains on corpora filtered at $40\%$ sequence identity against the benchmark test sets they most resemble, and is applied to ESM-2 (35M and 150M parameters) and ESM-C (300M).
%Results
Post-training makes structural family membership recoverable from the frozen embedding without a trained readout: on SCOPe-40, the adjusted Rand index of the 35M model rises from $0.054$ (near chance) to $0.507$, and family retrieval gains $+0.185$, $+0.190$ and $+0.424$ Recall@1 over the three frozen backbones, with paired bootstrap intervals excluding zero. On the CATH midnight-zone benchmark, homologous-superfamily accuracy rises from $40.7\%$ to $56.7\%$ at 35M and from $43.3\%$ to $62.7\%$ at 150M, more than the $12$ points a CATH-supervised model gains over its own base. Against a maximally sensitive HMMER search, alignment keeps the best single hit at both ESM-2 scales, but the embedding leads at ranking depth (MAP $0.705$ against $0.607$ at 150M) and returns a ranking for every query, including those on which alignment returns no hit. Under a trained linear probe, however, the same embeddings fall below the untuned backbone, as do structure-distilled embeddings against theirs, so the gain in neighborhood structure comes at a cost in linear-probe accuracy. Code and models are available at \url{https://github.com/oriel9p/ProtSent} and \url{https://huggingface.co/collections/oriel9p/protsent}.
\end{abstract}

%% ============================================================================
\section{Introduction}
\label{sec:intro}
%% ============================================================================

Protein language models (pLMs) learn per-residue representations by masked-language modeling over large sequence databases~\citep{rives2021biological, lin2023evolutionary, elnaggar2022prottrans}. Downstream use is usually mediated by a \emph{sequence-level} vector obtained by mean-pooling those residues, and nothing in the pretraining objective asks that vector to place biologically related proteins near one another. The geometry that retrieval, clustering and nearest-neighbor annotation transfer depend on is therefore never optimized. In natural language processing the same limitation was addressed by Sentence-BERT~\citep{reimers2019sentence}, whose contrastive objective directly optimizes embedding proximity, the quantity retrieval relies on.

We apply this principle to protein language models. ProtSent fine-tunes a pLM backbone end-to-end with Multiple Negatives Ranking Loss (MNRL)~\citep{henderson2017efficient} over three sources of biological relatedness, Pfam family membership~\citep{mistry2021pfam}, AlphaFold~DB Foldseek structural clusters~\citep{varadi2022alphafold, vankempen2024foldseek} and STRING interaction pairs~\citep{szklarczyk2023string}, with deep mutational scanning (DMS) fitness ordering~\citep{notin2023proteingym} under a CoSENT loss~\citep{su2022cosent} as an auxiliary fourth. Existing structure-informed sequence models distill a single type of relation from a single teacher; ProtSent combines family, structure and interaction as separate signals in one objective. We test whether contrastive post-training makes biologically meaningful structure recoverable from a frozen sequence-only embedding without a trained readout, using clustering statistics and retrieval on SCOPe-40 and CATH, alignment baselines, and paired bootstrap intervals throughout. Under a trained linear probe on the same embeddings, however, ProtSent performs worse than its own untuned backbone on most tasks.

Our contributions are the framework itself, applied at three model sizes (35M, 150M and 300M parameters) and across two model families, with its contrastive corpora filtered against the benchmark test sets they most resemble; a geometry result measured on the embedding space itself, with a matching retrieval gain on an external benchmark (CATH); the linear-probe result above, in which ProtSent trails its own backbone, as structure distillation also does; and ablations in which removing or adding one supervision source moves mainly the tasks related to it, most clearly STRING and PPI. Code for corpus construction, training and evaluation is available at \url{https://github.com/oriel9p/ProtSent}, and the trained models at \url{https://huggingface.co/collections/oriel9p/protsent}.

%% ============================================================================
\section{Related work}
\label{sec:related}
%% ============================================================================

\paragraph{Protein language models and contrastive fine-tuning.}
Self-supervised pLMs learn residue-level representations from large sequence databases: ESM-1b and ESM-2~\citep{rives2021biological, lin2023evolutionary} on UniRef~\citep{suzek2015uniref}, ProtTrans~\citep{elnaggar2022prottrans} at billion-parameter scale, with TAPE~\citep{rao2019evaluating} as a standardized benchmark. Convolutional alternatives remain competitive at matched pretraining~\citep{yang_convolutions_2024}, and bilingual sequence--structure models~\citep{heinzinger_bilingual_2024} sit between the sequence-only and structure-aware families. In all of these pLMs, transfer to a downstream task is mediated by a probe trained on top of the representation, and the arrangement of the sequence-level embedding space itself is never a training target. Sentence-BERT~\citep{reimers2019sentence, reimers2020making} showed in natural language that a siamese contrastive objective makes cosine similarity track semantic similarity. For proteins, ProtTucker~\citep{heinzinger2022contrastive} fine-tunes ProtT5 with a triplet loss on CATH superfamily labels, and \citet{redl2023optimizing} adapted SentenceTransformers to ProtBERT using disorder and stability annotations; both supervise a single label hierarchy or annotation type. ProtTucker is evaluated by nearest-neighbor annotation transfer on CATH, the protocol we reuse in Section~\ref{sec:results:cath}, and \citet{redl2023optimizing} by downstream predictors.

\paragraph{Structure-informed sequence models.}
A growing family of sequence-only pLMs injects structure at training time and consumes none at inference: ESM-S~\citep{zhang2024esms}, S-PLM~\citep{wang2025splm}, ISM~\citep{ouyangzhang2025ism} and Magneton~\citep{calef2026magneton}. All distill a structure encoder or structural tokens, so each supervises one kind of relation from one teacher. ProTrek~\citep{su2025protrek} adds text and structure as query modalities, and therefore needs them at query time.

\paragraph{Search, retrieval and function prediction.}
Traditional search relies on local alignment and profile HMMs~\citep{Soding2005, steinegger_mmseqs2_2017, eddy2011hmmer}. PLMSearch~\citep{liu2024plmsearch} re-ranks homology hits over ESM-1b embeddings, DHR~\citep{hong2024dense} trains a dual encoder on SCOPe domains, and Foldseek~\citep{vankempen2024foldseek} encodes backbone geometry into a structural alphabet, which requires a structure at query time. For function prediction, GOBeacon~\citep{lin2025gobeacon} applies contrastive regularization to multi-label GO classifiers~\citep{Jiang2016}, and unsupervised approaches do not necessarily beat supervised ones~\citep{michael-pitschaze_detecting_2024, ofer_protein_2025, ofer_language_2021}. ColBERT~\citep{khattab2020colbert} scores a pair of texts by per-token matching (MaxSim) instead of by comparing two pooled vectors.

\paragraph{Relation to prior work.}
ProtSent differs from the work above in three respects. It supervises several types of relation together (family, structural cluster and interaction, with fitness as an auxiliary signal), whereas ProtTucker and the structure-distillation models supervise one and DHR trains on curated SCOPe domains; it requires neither structure nor text at query time, unlike Foldseek and ProTrek; and it evaluates the resulting embedding space directly, with clustering and retrieval statistics, in addition to a downstream probe. We do not claim to outperform the structure-distillation models in general, since we ran only one of them (ISM-C-300M, Section~\ref{sec:results:probes}).

%% ============================================================================
\section{Method}
\label{sec:method}
%% ============================================================================

\subsection{ProtSent formulation and backbones}
\label{sec:method:formulation}

Let $f_\theta: \mathcal{S} \rightarrow \mathbb{R}^d$ be a function with parameters $\theta$, the weights of a pLM backbone, mapping an amino-acid sequence to a $d$-dimensional embedding, where $d$ is the backbone's hidden dimension. Given pairs of amino-acid sequences $(s_i, s_j)$, we want to fine-tune $\theta$ so that the similarity of the embedded sequences, $\text{sim}(f_\theta(s_i), f_\theta(s_j))$, is high for related pairs and low for unrelated ones. After fine-tuning the encoder is frozen, and every result that follows uses its frozen embeddings.

We use ESM-2~\citep{lin2023evolutionary} at 35M ($d{=}480$) and 150M ($d{=}640$), and ESM-C 300M~\citep[$d{=}960$;][]{esmc2024} as a second, independently pretrained model family, each wrapped in SentenceTransformers~\citep{reimers2019sentence}. All Transformer layers are updated. Sequence embeddings are the mean over non-padding residue tokens. At evaluation, sequences longer than 1{,}024 residues are truncated to their first 1{,}024 and shorter sequences are used in full. Per-scale configurations and wall clock are in Appendix~\ref{app:training}, which also discloses a training-time defect: the requested 512-residue training cap was not applied, so the primary ESM-2 models trained on untruncated sequences.

\subsection{Training data}
\label{sec:method:data}

Positives are Pfam family co-membership~\citep{mistry2021pfam}; Foldseek structural-cluster co-membership over AlphaFold~DB sequences at pLDDT $>70$, the AlphaFold~DB threshold for a confident prediction~\citep{varadi2022alphafold, barrio-hernandez_clustering_2023, vankempen2024foldseek}, which are unsupervised clusters of predicted structures, as opposed to curated SCOP families; and STRING v12 physical interaction at \texttt{combined\_score} $\geq 400$, STRING's medium-confidence cut-off~\citep{szklarczyk2023string}, a relation distinct from family and structural similarity. The auxiliary DMS source is ProteinGym~\citep{notin2023proteingym}, where each row is a pair: (wild type, mutant, $s$), with $s$ the per-assay $z$-score clipped to $[-3,3]$ and rescaled to $[0,1]$. Assays whose identifiers name our benchmarks are dropped, and Appendix~\ref{app:datasets} measures how much overlap that leaves: none for GB1, whose assays the rule names, and $2.2\%$ of the corpus for the fluorescence test set, whose related assays it does not.

We also generated synthetic Pfam hard negatives and do not use them: an ablation prefers the model without them (Section~\ref{sec:ablations}), and an HMMER audit found that 543 of 543 sampled mutants still matched the family they were meant to be negatives for (Appendix~\ref{app:datasets}).

\subsection{Decontamination and verification}
\label{sec:method:decontam}

We filter all three MNRL corpora with MMseqs2 \texttt{easy-search}, corpus as query and benchmark test set as target, at $40\%$ identity and $80\%$ coverage, dropping every corpus sequence with a hit: Pfam $28.53\mathrm{M} \rightarrow 27.93\mathrm{M}$ ($2.11\%$ removed), AlphaFold~DB $135.40\mathrm{M} \rightarrow 126.30\mathrm{M}$ ($6.72\%$), STRING $76.07\mathrm{M} \rightarrow 71.89\mathrm{M}$ ($5.49\%$). Pfam and AlphaFold~DB were filtered against the remote-homology test set (3{,}244 sequences) and STRING against the Bernett PPI test set (3{,}022 sequences); the other benchmark test sets were not used as filter targets, which limits the scope of this control. We then checked the filtering itself, by joining the written training files back against the recorded removal lists: no flagged sequence survives in any corpus.

Two further limits apply to this decontamination. SCOPe-40 has no train/test split to filter, and the median maximum identity of a SCOPe domain to our corpus is $0.908$, so a corpus-level filter would remove close relatives of most SCOPe-40 domains from training; we therefore filter the \emph{benchmark} instead (Section~\ref{sec:results:decontam}). And no identity threshold excludes fold-level overlap, since supervision is Foldseek-cluster and Pfam-family co-membership, so a training pair sharing a query's fold at $15\%$ identity survives any filter. We did not run a fold-exclusion control, so our structural benchmarks measure retrieval within the fold space covered by the AlphaFold~DB clusters, not generalization to unseen folds.

\subsection{Training objective}
\label{sec:method:objective}

The primary objective is MNRL~\citep{henderson2017efficient} over a batch of $N$ anchor--positive pairs $\{(s_i, s_i^+)\}_{i=1}^N$, treating the other positives in the batch as negatives:

\begin{equation}
  \mathcal{L}_{\text{MNRL}} = -\frac{1}{N} \sum_{i=1}^{N} \log \frac{\exp(\text{sim}(f_\theta(s_i), f_\theta(s_i^+)) / \tau)}{\sum_{j=1}^{N} \exp(\text{sim}(f_\theta(s_i), f_\theta(s_j^+)) / \tau)},
  \label{eq:mnrl}
\end{equation}

where $s_i^+$ is the positive paired with anchor $i$, the denominator ranges over the positives of all $N$ pairs, $\text{sim}$ is cosine similarity, and $\tau = 0.05$ is the temperature: the SentenceTransformers default (a scale of 20), which we did not tune, and which sharpens the softmax so that the hardest in-batch negatives dominate the gradient.

\paragraph{In-batch negatives.} An optimizer batch formed by gradient accumulation does not share in-batch negatives across accumulation steps, so the negative pool of each MNRL call is only the per-device batch (Appendix~\ref{app:training}). The primary models therefore use a cached MNRL implementation with a true 1{,}024-example batch per device, so $N = 1024$ in Equation~\ref{eq:mnrl}.

\paragraph{Auxiliary DMS objective.} CoSENT~\citep{su2022cosent} is ordinal over pairs: within a batch, if pair $p$ scores above pair $q$, the loss pushes the wild-type-to-mutant cosine of $p$ above that of $q$. There is no absolute cosine target, so the objective does not flatten an assay. The biological assumption also has a weakness: $s$ is a normalized fitness score, so mutants are ordered by absolute fitness instead of by their deviation from wild-type fitness. A variant whose fitness exceeds the wild type is therefore placed closer to it than one whose fitness matches it, which reverses the intended ordering. Using $|s - s_{\text{WT}}|$ as the key would correct this; we did not retrain with it. The pairing is additionally wild-type-anchored, so mutant--mutant geometry is constrained only indirectly. We therefore keep DMS strictly auxiliary: the primary ESM-2 models train without it.

\paragraph{Sampling.} Batches are drawn from one source per step, with sources sampled in proportion to their size; round-robin sampling performed slightly worse in our ablations (Section~\ref{sec:ablations}).

%% ============================================================================
\section{Experiments}
\label{sec:experiments}
%% ============================================================================

\textbf{ProtSent} denotes the models trained on the filtered corpora with the configuration our ablations favor, at 35M and 150M parameters on ESM-2 and, as \textbf{ProtSent-ESMC-300M}, with the same recipe on ESM-C plus the auxiliary DMS source (Appendix~\ref{app:training}). Two further configurations appear where needed: the \textbf{initial configuration}, trained on unfiltered corpora with round-robin sampling and synthetic hard negatives, on which the single-factor ablations of Section~\ref{sec:ablations} were run; and \textbf{ProtSent+DMS}, a continuation of ProtSent with the DMS target added, used for the source-specificity experiment. For every ProtSent model we report its own frozen backbone, since this is the only comparison that controls for pretraining.

\looseness=-1 All encoders are frozen, and we report two probes on identical splits and never average them: \textbf{3-NN} ($k{=}3$, uniform, Euclidean), which measures neighborhood quality, and \textbf{linear} (logistic regression or ridge on standardized features), which corresponds to typical practice. Full probe settings are in Appendix~\ref{app:protocol}. The suite spans 23 tasks drawn from TAPE, FLIP, PEER and related sources~\citep{rao2019evaluating, dallago2021flip, xu2022peer, khurana2018deepsol, teufel2022signalp, ofer2014profet}; SCOPe-40 retrieval enters as one task among the 23 and is also reported on its own in Section~\ref{sec:results:scope}. Two tasks sit outside the sweep because they need a different input protocol and are reported separately: Bernett PPI~\citep{bernett2024ppi}, which is pair-input (each partner embedded independently and the two vectors concatenated; Section~\ref{sec:ablations}), and RhlA enzyme mutations, whose sequences are 6-residue mutation-site strings (Section~\ref{sec:discussion}). Three tasks (remote homology, antibiotic resistance, temperature stability) have an undefined one-vs-rest AUC because the test split contains classes absent from training, so every win/tie/loss tally covers the 20 tasks with a defined main metric at a tie band of $0.005$, with those three reported separately by accuracy (Appendix~\ref{app:protocol}). Per-task tables are in Appendix~\ref{app:pertask}.

Retrieval metrics are per-query means, so retrieval comparisons use paired bootstrap resamples over the same queries scoring every method, $10{,}000$ for the SCOPe-40 contrasts and $5{,}000$ for the late-interaction contrasts of Appendix~\ref{app:late}. The 23-task suite has one measurement per cell and no intervals; we report those values as measurements and draw no inferential claim from the aggregate, since a sign test over the tallies resolves only our own linear-probe losses.

\looseness=-1 \paragraph{Retrieval benchmarks.} SCOPe-40 is an all-versus-all leave-one-out retrieval over 2{,}207 domains~\citep{fox2014scope}, scored at family level. Only the 1{,}693 queries with a non-self same-family protein in the gallery can succeed at any $k$, and all figures are means over those. CATH v4.3 midnight-zone transfer follows \citet{heinzinger2022contrastive} exactly: move a CATH~\citep{sillitoe2021cath} label from 69{,}605 lookup domains to 219 queries filtered so that no alignment-detectable relative exists in the lookup set, by 1-NN over mean-pooled embeddings.

\paragraph{Alignment baselines.} Both tools were run over the whole benchmark through the same scoring code as the embeddings, with no-hit queries scored as failures. MMseqs2~\citep{steinegger_mmseqs2_2017} runs at high sensitivity (\texttt{-s 7.5 -e 10}); its default \texttt{-s 5.7} is much weaker (SCOPe-40 R@1 $0.385$). HMMER phmmer~\citep{eddy2011hmmer} is reported in its filters-off, maximally sensitive configuration; a default-filter run gives R@1 $0.697$ against $0.753$, which would overstate our models.

%% ============================================================================
\section{Results}
\label{sec:results}
%% ============================================================================

\subsection{Embedding-space geometry}
\label{sec:results:geometry}

\begin{table}[t]
  \caption{Embedding-space geometry of the frozen mean-pooled embeddings on SCOPe-40 (2{,}207 domains, 917 families), ESM-2 35M line. ARI, NMI and silhouette are computed by clustering at the true number of families; P@10 is precision at 10 over the 1{,}693 eligible queries (attainable ceiling $0.516$); the last five columns give the mean cosine distance between domain pairs sharing 0 to 4 levels of the SCOPe hierarchy, which should decrease monotonically if the hierarchy is encoded. Figure~\ref{fig:hierarchy} in Appendix~\ref{app:geometry} plots the distance profiles.}
  \label{tab:geometry}
  \centering
  \footnotesize
  \setlength{\tabcolsep}{4pt}
  \begin{tabular}{lcccc ccccc}
    \toprule
    & & & & & \multicolumn{5}{c}{Mean distance by shared SCOPe levels} \\
    \cmidrule(lr){6-10}
    Model & ARI & NMI & Silhouette & P@10 & 0 & 1 & 2 & 3 & 4 \\
    \midrule
    ESM-2 35M (frozen base) & 0.054 & 0.823 & $-0.143$ & 0.279 & 0.156 & 0.140 & 0.146 & 0.123 & 0.064 \\
    ProtSent 35M            & 0.507 & 0.917 & $+0.053$ & 0.367 & 0.865 & 0.821 & 0.780 & 0.571 & 0.299 \\
    \bottomrule
  \end{tabular}
\end{table}

Table~\ref{tab:geometry} reports clustering statistics over the frozen mean-pooled embeddings of the 2{,}207 SCOPe-40 domains at their 917 families, together with the mean pairwise distance as a function of shared hierarchy depth. The ARI of the untuned 35M model is close to chance; post-training raises it to $0.507$ and moves the silhouette from negative to slightly positive, so families go from overlapping to marginally separated. The distance profile is a more informative test, since a representation that encodes the hierarchy should give distances that decrease monotonically with shared depth. ProtSent does; for ESM-2 35M, domains sharing two levels sit further apart than domains sharing one. In the untuned space, family membership is encoded weakly and non-monotonically; after post-training it is encoded monotonically, over a roughly six-fold wider range of distances ($0.57$ against $0.09$ between unrelated pairs and pairs sharing all four levels).

\subsection{SCOPe-40 retrieval}
\label{sec:results:scope}

\begin{table}[t]
  \caption{SCOPe-40 family-level retrieval, leave-one-out over the 1{,}693 eligible queries of 2{,}207; self excluded, no-hit scored as failure. ESM-C rows differ from the ESM-2 rows in both model family and scale and are not a controlled comparison; only within-backbone differences are interpretable. The ESM-C-line intervals come from a re-encode reproducing these rows to within $0.002$ (Appendix~\ref{app:structural}).}
  \label{tab:scope}
  \centering
  \footnotesize
  \setlength{\tabcolsep}{4pt}
  \begin{tabular}{llccc}
    \toprule
    & Method & Recall@1 & Recall@10 & MAP \\
    \midrule
    \multirow{2}{*}{\textit{Alignment}}
      & MMseqs2 (\texttt{-s 7.5}) & 0.656 & 0.740 & 0.410 \\
      & HMMER phmmer (filters off) & 0.753 & 0.898 & 0.607 \\
    \midrule
    \multirow{2}{*}{\textit{ESM-2 35M}}
      & ESM-2 35M (frozen base)   & 0.499 & 0.761 & 0.421 \\
      & ProtSent 35M              & 0.685 & 0.922 & 0.646 \\
    \midrule
    \multirow{2}{*}{\textit{ESM-2 150M}}
      & ESM-2 150M (frozen base)  & 0.554 & 0.770 & 0.424 \\
      & ProtSent 150M             & 0.743 & 0.937 & 0.705 \\
    \midrule
    \multirow{3}{*}{\textit{ESM-C 300M}}
      & ESM-C 300M (frozen base)  & 0.371 & 0.579 & 0.221 \\
      & ISM-C 300M (structure-distilled) & 0.430 & 0.659 & 0.273 \\
      & ProtSent-ESMC-300M        & \textbf{0.797} & \textbf{0.954} & \textbf{0.769} \\
    \bottomrule
  \end{tabular}
\end{table}

\paragraph{Gains over the frozen backbones.} Over the 1{,}693 eligible queries, ProtSent minus its base is $+0.185\,[+0.162,+0.210]$ Recall@1 and $+0.225\,[+0.210,+0.240]$ MAP at 35M, and $+0.190\,[+0.165,+0.214]$ and $+0.281\,[+0.265,+0.297]$ at 150M. % CHECK: MAP intervals now taken from Table 23 (5,000 resamples), whose point estimates equal the observed differences; the R@1 intervals are from the 10,000-resample run.
 Structure distillation moves its own backbone by $+0.060\,[+0.034,+0.085]$ Recall@1 and $+0.053\,[+0.038,+0.067]$ MAP (ISM-C minus ESM-C). Applying the recipe to ESM-C, with the DMS source added, gives the largest within-backbone gain in the paper, $+0.424\,[+0.397,+0.451]$ Recall@1 and $+0.548\,[+0.532,+0.565]$ MAP, and places both post-training methods on the same backbone; part of this margin reflects the weak starting point of raw ESM-C for nearest-neighbor retrieval.

\looseness=-1 \paragraph{Comparison with alignment.} The embedding advantage is in ranking depth: MAP $0.705$ against $0.607$ at 150M, Recall@10 $0.937$ against $0.898$; over the frozen backbone, precision@10 also rises from $0.279$ to $0.367$ at 35M (Table~\ref{tab:geometry}). % TODO(phmmer): 691 and the +0.017 below come from the DEFAULT-filter phmmer run (the one with R@1 0.697), not the filters-off run of Table 2. Re-run scripts/phmmer_scope40_nohit.py and replace with the filters-off counts over all 2,207 and over the 1,693 eligible queries.
The embedding also returns a ranking for every query, whereas phmmer at its default filters returns no hit for 691 of the 2{,}207 queries, which then score zero at every cutoff. Alignment retains the higher top-1 recall at both ESM-2 scales ($0.753$ against $0.685$ and $0.743$); ProtSent-ESMC is the only model above it ($0.797$), a comparison we report without a paired test because per-query phmmer records were not stored. Part of the margin at depth is due to this coverage gap: default-filter phmmer and MMseqs2 both flatten between Recall@10 and Recall@30 ($+0.017$) whereas the embeddings do not ($+0.073$, $+0.041$).

\subsection{Sensitivity to sequence identity with the training corpus}
\label{sec:results:decontam}

SCOPe-40 cannot be filtered at the corpus level (Section~\ref{sec:method:decontam}), so we filter the benchmark instead. Restricting the evaluation to queries below a given maximum identity to our corpus does not change the conclusion (Appendix~\ref{app:structural}): on the 164 queries with the lowest identity to the training corpus, ProtSent 35M still leads phmmer at depth ($+0.116\,[+0.049,+0.189]$ Recall@10, $+0.140\,[+0.075,+0.207]$ MAP) and is unresolved at top-1 ($-0.043\,[-0.128,+0.043]$); this analysis is paired against default-filter phmmer (Table~\ref{tab:clean}) and bounds the effect of sequence identity only. The per-query correlation between identity to the corpus and gain in average precision is \emph{negative} at 35M ($\rho = -0.116$, $p < 10^{-3}$) and null at 150M after controlling for baseline headroom. Memorization of training sequences would predict a positive correlation, which we do not observe at either scale.

On remote homology, the task Pfam and AlphaFold~DB were filtered against, the outcome depends on the probe. Under 3-NN, ProtSent gains at both scales (accuracy $0.584 \rightarrow 0.667$ at 35M and $0.519 \rightarrow 0.661$ at 150M). Under a linear probe at 150M, accuracy is level with the backbone ($0.750$ for both) and macro-F1 falls ($0.516 \rightarrow 0.494$), a deficit that sits mostly (about $86\%$) in classes with two or fewer test examples (Appendix~\ref{app:geometry}). The initial configuration, trained on the unfiltered corpora, scores higher on this task at 150M ($0.705$ under 3-NN), which is consistent with the larger unfiltered model having benefited from leakage, although the two runs also differ in other settings (Appendix~\ref{app:structural}).

\subsection{Linear-probe evaluation}
\label{sec:results:probes}

\begin{table}[t]
  \caption{Win/tie/loss of each post-trained model against \emph{its own} frozen backbone, over the 20 tasks with a defined main metric, tie band $0.005$, with the median signed delta. These tallies do not support comparative claims between models; a sign test is significant only for the linear-probe record marked $^{*}$. EC, GO-MF and SCOPe-40 are probe-invariant (Appendix~\ref{app:protocol}); excluding them, the four linear-probe records are 1W/6T/10L, 3W/3T/11L, 3W/2T/12L and 6W/3T/8L.}
  \label{tab:probes}
  \centering
  \footnotesize
  \setlength{\tabcolsep}{4pt}
  \begin{tabular}{llcc}
    \toprule
    Model & Backbone & 3-NN probe & Linear probe \\
    \midrule
    ProtSent 35M        & ESM-2 35M   & 10W/3T/7L\ \ $+0.004$ & 2W/7T/11L\ \ $-0.011^{*}$ \\
    ProtSent 150M       & ESM-2 150M  & 10W/3T/7L\ \ $+0.005$ & 4W/4T/12L\ \ $-0.013$ \\
    ProtSent-ESMC-300M  & ESM-C 300M  & 14W/0T/6L\ \ $+0.013$ & 6W/2T/12L\ \ $-0.013$ \\
    \midrule
    ISM-C 300M (structure distillation) & ESM-C 300M & 7W/1T/12L\ \ $-0.008$ & 7W/3T/10L\ \ $-0.006$ \\
    \bottomrule
  \end{tabular}
\end{table}

Table~\ref{tab:probes} reports the main negative result of the paper. Under a 3-NN probe, which depends on local neighborhood structure, contrastive post-training is net positive at every scale and in both model families; the two ESM-2 scales behave similarly (10W/3T/7L each), so the effect on this suite does not grow with backbone size, whereas the retrieval MAP and CATH gains do. Under a linear probe, which measures what a trained readout can recover, it is net negative everywhere, and this is the only direction in which a sign test is significant. We therefore do not claim that ProtSent yields better general-purpose embeddings. Compressing family, structural-cluster and interaction relations into a single metric improves neighborhood structure but reduces the property information that a linear readout can recover.

\looseness=-1 Two further observations bear on this cost. First, structure distillation shows a similar one. ISM-C-300M~\citep{ouyangzhang2025ism} is distilled from ESM-C-300M, so vanilla ESM-C is its matched control (identical architecture, parameter count and tokenizer), and its record against that base is in Table~\ref{tab:probes}. Under 3-NN ProtSent is net positive where ISM-C is net negative, and under a linear probe both are net negative (Fisher exact test on the contrast, $p = 0.22$). The two methods differ in which tasks they affect: under 3-NN, ISM-C gains on cloning, fluorescence and solubility ($+0.180$, $+0.173$, $+0.091$) and loses on stability, EC and GB1 ($-0.114$, $-0.113$, $-0.079$), whereas our supervision sources affect a different set of tasks (Section~\ref{sec:ablations}). Second, the deficit does not come from a collapse of the space. Post-training moves it toward isotropy (mean random-pair cosine $0.848$ for ESM-2 35M against $0.152$ for ProtSent) and raises its participation ratio from $7.9$ to $52.5$ of 480 dimensions (Appendix~\ref{app:geometry}), so post-training spreads the pooled embedding over more directions. Section~\ref{sec:results:late} examines the readout instead: a residue-level readout on the same frozen weights, without training, recovers much of what mean pooling discards.

\subsection{CATH midnight-zone benchmark}
\label{sec:results:cath}

\begin{table}[t]
  \caption{CATH v4.3 midnight-zone annotation transfer, accuracy (\%) per level, 1-NN, lookup-69k $\rightarrow$ test-219, on the published splits of \citet{heinzinger2022contrastive}. Our MMseqs2 row, which reproduces their published $35$, serves as a check that our pipeline follows their protocol; the like-for-like comparison is each model against its own frozen base. $95\%$ CIs are $\pm 5$ at C and $\pm 8$ at H ($n{=}150$). For reference, \citet{heinzinger2022contrastive} report, with their own models and code, C/A/T/H of 84/67/57/64 for raw ProtT5, 89/75/64/76 for ProtTucker(ProtT5) and 70/60/59/77 for profile HMMER. Full arms and controls in Appendix~\ref{app:structural}.}
  \label{tab:cath}
  \centering
  \footnotesize
  \setlength{\tabcolsep}{4pt}
  \begin{tabular}{lrrrr}
    \toprule
    Model & C & A & T & H \\
    \midrule
    ESM-2 35M (frozen base)   & 78.5 & 54.3 & 42.4 & 40.7 \\
    \textbf{ProtSent 35M}     & 82.2 & 64.4 & 53.3 & \textbf{56.7} \\
    ESM-2 150M (frozen base)  & 74.0 & 53.0 & 41.0 & 43.3 \\
    \textbf{ProtSent 150M}    & 84.0 & 69.9 & 57.1 & \textbf{62.7} \\
    ESM-C 300M (frozen base)  & 66.7 & 36.1 & 21.9 & 18.7 \\
    ISM-C 300M                & 80.8 & 47.9 & 29.5 & 25.3 \\
    MMseqs2 (ours / published $35$) & -- & -- & -- & 34.7 \\
    \bottomrule
  \end{tabular}
\end{table}

\looseness=-1 Each ProtSent arm gains $+3.7$, $+10.1$, $+10.9$ and $+16.0$ over its own base at 35M, and $+10.0$, $+16.9$, $+16.1$ and $+19.4$ at 150M, against ProtTucker's published $+5$/$+8$/$+7$/$+12$ over ProtT5. The pattern is similar, with larger gains at the finer levels; structure distillation shows the same direction at smaller magnitude ($+6.6$ at H). In absolute terms, ProtSent-150M is close to the published raw-ProtT5 numbers at every level with a much smaller model. We interpret these results as evidence for the pattern and for the size of the two H-level gains; we do not rank models against adjacent rows, and profile HMMs still lead at the finest level ($77$ against $62.7$). Three controls in Appendix~\ref{app:structural} address the main alternative explanations: direction-only normalization of the vanilla embeddings adds at most $+2.7$ points, ESM-2 is flat with scale on this task ($40.7 \rightarrow 43.3 \rightarrow 42.7$ from 35M to 650M), and a 3-mer baseline scores $0.00$. Three factors bound the comparison: the queries have no alignment-detectable relative in the lookup set by construction; ProtTucker is supervised on CATH labels, whereas our supervision never uses a CATH class; and the base and the ProtSent model share identical pretraining exposure, so the difference between them is attributable to the contrastive stage.

\subsection{Ablation of supervision sources}
\label{sec:ablations}

\begin{table}[t]
  \caption{Single-factor ablations on the initial configuration (Section~\ref{sec:experiments}; ESM-2 35M, default split, single run). Each cell is the relative change (\%) of the task metric against stock ESM-2 35M when one source or setting is changed; the full 23-task table is Appendix~\ref{app:ablation_full}. These are single-run values and support only the direction of source-specific effects.}
  \label{tab:ablation}
  \centering
  \footnotesize
  \setlength{\tabcolsep}{3pt}
  \begin{tabular}{lcc rrrrr}
    \toprule
    Configuration & Tasks improved & Mean $\Delta$\% & Remote hom. & EC & PPI & Fluorescence & GB1 \\
    \midrule
    Full (initial config.) & 16/23 & $+6.7$ & $+40.5$ & $+0.3$  & $+5.3$ & $+15.6$ & $-0.8$ \\
    $-$ Pfam               & 15/23 & $+4.6$ & $+42.5$ & $-1.3$  & $+5.9$ & $+20.6$ & $+9.0$ \\
    $-$ AlphaFold~DB       & 13/23 & $+3.2$ & $+15.3$ & $-11.1$ & $+5.5$ & $+22.0$ & $+11.8$ \\
    $-$ STRING             & 17/23 & $+5.9$ & $+43.3$ & $+3.7$  & $-0.5$ & $+19.6$ & $+13.8$ \\
    $-$ DMS                & 15/23 & $+5.8$ & $+40.6$ & $-2.1$  & $+5.9$ & $+10.4$ & $+14.7$ \\
    $-$ Hard negatives     & 20/23 & $+7.9$ & $+49.5$ & $+7.7$  & $+4.3$ & $+24.2$ & $+7.9$ \\
    Proportional sampling  & 16/23 & $+7.0$ & $+54.7$ & $+2.5$  & $+6.6$ & $+18.4$ & $+8.4$ \\
    \bottomrule
  \end{tabular}
\end{table}

Table~\ref{tab:ablation} summarizes single-factor ablations on the initial configuration. Structural supervision is the largest single contributor: removing AlphaFold~DB reduces the number of improved tasks from 16 to 13 and the mean gain from $+6.7\%$ to $+3.2\%$, and it is the only removal that costs remote homology and the one that costs enzyme-class prediction most, whereas removing Pfam costs less (15 tasks, $+4.6\%$). Removing STRING is the only change that turns the PPI gain negative. The fitness columns are less clean: removing DMS lowers the fluorescence gain by a third, but fluorescence and GB1 also rise when other sources are removed. The initial configuration's own defaults are not supported by the ablation either: dropping synthetic hard negatives, which the audit of Section~\ref{sec:method:data} shows were in effect noisy positives, improves 20 of 23 tasks at $+7.9\%$, and proportional sampling edges round-robin ($+7.0\%$ against $+6.7\%$), so the final configuration uses neither. The GB1 column runs against the DMS interpretation, since removing DMS improves it; we return to this in Appendix~\ref{app:ablation_full}.

Because these are single-run relative changes on the initial configuration, we add a second experiment on the final one. ProtSent+DMS continues ProtSent with the DMS target added. At 35M, under a linear probe the gains are concentrated on the DMS-derived tasks (stability $+0.095$, AAV $+0.073$, fluorescence $+0.015$, GB1 $+0.013$, beta-lactamase $+0.011$) while the suite medians move by $+0.001$ under both probes. At 150M the same tasks move under the 3-NN probe (fluorescence $+0.068$, GB1 $+0.025$, beta-lactamase $+0.021$), but under a linear probe stability and AAV fall (Table~\ref{app:tab:150m_lin}), so the source-specific interpretation rests on the 35M results; fluorescence, stability and beta-lactamase also overlap the DMS corpus (Table~\ref{app:tab:dms_overlap}), so their gains cannot be separated from that overlap. ProtSent+DMS changes several settings at once relative to ProtSent (Appendix~\ref{app:training}); a GOR-off ablation rules out that term, but the DMS target remains confounded with an annealed schedule, a 512-residue cap and a fresh data draw, so the change cannot be attributed to the DMS objective alone. Its overall record against the backbone is similar to ProtSent's (13W/2T/5L under 3-NN and 4W/2T/14L under a linear probe at 150M; Appendix~\ref{app:pertask}). The movement is concentrated on the tasks nearest to the added source, and it does not establish a general link between DMS training and fitness prediction. Interaction supervision likewise survives filtering: the $+5.3\%$ PPI figure above predates the STRING filter, and re-measured after $4{,}178{,}737$ pairs were removed, Bernett PPI AUC under a 3-NN probe is $0.572 \rightarrow 0.583$ at 35M and $0.572 \rightarrow 0.594$ at 150M. On this benchmark no method, including ours, separates the task under a linear probe, so this is a nearest-neighbor result only.

Few-shot results with absolute scores, a fixed estimator and five seeds are in Appendix~\ref{app:seeds}; on remote homology under a 3-NN probe, ProtSent's gain over the backbone grows with the label budget, from $+0.010$ at $N{=}100$ to $+0.104$ at $N{=}1000$ ($0.289$ against $0.185$); there is none at $N{=}50$ ($0.045$ against $0.061$).

\subsection{Late-interaction scoring of frozen embeddings}
\label{sec:results:late}

\looseness=-1 Scoring the same frozen weights with MaxSim, the late-interaction operator of ColBERT~\citep{khattab2020colbert}, instead of pooled cosine improves SCOPe-40 MAP for every model we tested, with paired intervals excluding zero: $+0.124$ and $+0.182$ for ESM-2 35M and 150M, $+0.038$ and $+0.045$ for ProtSent (Table~\ref{tab:late}, Appendix~\ref{app:late}). The gain is larger for the untuned backbones than for ProtSent, and post-training remains the dominant effect: under identical MaxSim scoring ProtSent leads its own backbone by $+0.139\,[+0.124,+0.154]$ MAP at 35M and $+0.144\,[+0.130,+0.158]$ at 150M, roughly three times what MaxSim adds to ProtSent. Substituting MaxSim for pooled distance inside the 3-NN probe, with splits, $k$, training set and metrics unchanged, improves 27 of 30 model-task pairs over pooled cosine and beats a trained linear probe on 19 of 30, with most of the losses on the two multiclass tasks, whose labels appear to be linearly separable in the pooled space. Per-residue embeddings therefore retain signal that mean pooling discards, and a residue-level readout recovers much of it without training; this does not show that the linear-probe cost of Section~\ref{sec:results:probes} is closed, since MaxSim helps the untuned backbone more (Table~\ref{tab:late}). A pooled shortlist re-ranked by MaxSim recovers $99.8\%$ of exhaustive MaxSim MAP at 100 candidates per query, so the stored index stays pooled-size (Appendix~\ref{app:late}).

%% ============================================================================
\section{Discussion}
\label{sec:discussion}
%% ============================================================================

\paragraph{Choice of method.}
Alignment is better at single-best-hit retrieval (Appendix~\ref{app:alignment}) and beats every 35M arm on 3 of 20 comparable tasks under a 3-NN probe and 6 under a linear probe at hit coverage above $0.90$, so when a query has alignable relatives and a single best hit is required, a sensitive alignment search is preferable and cheaper. The other side of the trade-off is coverage: phmmer at its default filters finds no hit for $47.6\%$ of the remote-homology test set and $31.3\%$ of the SCOPe-40 queries, and on RhlA, whose sequences are 6-residue mutation-site strings, coverage is $0.004$, so the reported score largely reflects the fallback behavior. An embedding always returns a ranked list, which is a practical advantage. On the same ESM-C backbone, ProtSent leads structure distillation on structural retrieval (Table~\ref{tab:scope}), while ISM-C does better on some property tasks (Section~\ref{sec:results:probes}); ProtSent can also combine family, structure and interaction, for which there is no single teacher to distill from. Adding fitness without losing variant sensitivity remains open (limitation (viii)). When the downstream step is a trained readout on frozen features, the untuned backbone is the better starting point.

\paragraph{Limitations.}
Five limitations bear directly on the central claim.
(i) Fold-level overlap is untested; no identity threshold excludes it, and neither SCOPe-40 nor CATH test-219 was filtered against our corpora at the fold level.
(ii) The 23-task aggregate does not support statistical inference (Section~\ref{sec:experiments}).
(iii) One training run exists per model, so training-seed variance is unmeasured; a near-trough checkpoint of each ProtSent run differs from the final one by up to $0.008$ at 35M and up to $0.021$ at 150M on the SCOPe-40 metrics (Appendix~\ref{app:structural}), so we treat no structural difference below $0.01$ at 35M or $0.02$ at 150M as resolved.
(iv) The final configuration was selected on these benchmarks: the ablations were scored on the same suite, a selection channel the corpus filter does not address, and one of their cells contradicts the DMS interpretation (GB1 in Table~\ref{tab:ablation}). SCOPe-40 is not among the ablation tasks (Table~\ref{tab:ablation_full}), and both alignment baselines were run after the configuration was fixed.
(v) The single-factor ablations were run on the initial configuration, which differs from the final one in corpus filtering, sampler, hard negatives and data budget at once; re-running them on the final configuration was beyond our compute budget.
\looseness=-1 Limitations (vi) to (x), in Appendix~\ref{app:limitations}, cover the unfiltered and measurably contaminated DMS corpus, the DMS objective's conflation of high fitness with wild-type-likeness, variant-effect scoring (where no embedding method is competitive), the unrun baselines (ESM-S, S-PLM, Magneton, Foldseek, PLMSearch, DHR, ProTrek, \citet{redl2023optimizing}, and a fine-tuning sweep), and the single-seed late-interaction runs.

% \paragraph{Future work.}
% \looseness=-1 If one latent space cannot hold family, structure, interaction and fitness at once without costing a trained readout, as our probes suggest, then per-stream tokens supervised separately, a multi-CLS extension of BERT, are the obvious alternative to one pooled vector.

\paragraph{Conclusion.}
\looseness=-1 Contrastive post-training on several types of biological relation, with no structure at inference, makes structural family membership recoverable from a frozen embedding without a trained readout, at three model sizes and in two model families, on filtered corpora and on an external benchmark. It does not make the backbone a better general-purpose encoder under a linear readout, and neither does structure distillation; a residue-level readout recovers part of what pooling discards. Alignment remains better at single-best-hit search, and the ablations suggest that each supervision source mainly moves the tasks related to it.
\label{sec:conclusion}

\label{BODY-END}
\subsubsection*{Reproducibility statement}
Section~\ref{sec:method} and Appendix~\ref{app:training} give the training configuration of every model (backbone, loss, batch and in-batch negative pool, sampler, steps, and hardware). Appendix~\ref{app:datasets} gives the construction of every training corpus with the exact filters, thresholds and seeds, the decontamination procedure of Section~\ref{sec:method:decontam} and its verification, and the measured overlap of the DMS corpus with its benchmarks. Appendix~\ref{app:protocol} gives the probe settings, and Appendices~\ref{app:pertask} to~\ref{app:ablation_full} give per-task results for every model and probe, the alignment-baseline configurations, the structural-benchmark controls, the seed study and the full ablation tables. Code for corpus construction, training and the evaluation suite is available at \url{https://github.com/oriel9p/ProtSent}, and the trained models at \url{https://huggingface.co/collections/oriel9p/protsent}.

\subsubsection*{Ethics statement}
This work uses only publicly available protein sequence, structure and interaction databases and published benchmarks, and involves no human subjects or personal data. The released models are general-purpose protein embedding models; we are not aware of misuse risks specific to them beyond those of the underlying protein language models.

\subsubsection*{AI use statement}
In this work, we used generative AI tools for writing, the tools were used for editing of the manuscript text, namely rephrasing sentences and checking grammar to improve readability. For software, they provided coding assistance for some of the scripts used to run our experiments. We did not use generative AI tools to generate synthetic data, develop theoretical models or conceptual frameworks, formulate or prove mathematical claims, propose or refine hypotheses, design the research methodology or experiments, clean or reformat datasets, perform translation, conduct data analysis, or interpret results. All AI-suggested edits to the text were reviewed by the authors, who wrote the original content and verified that the edits preserved the intended meaning. All AI-assisted code was reviewed, tested, and verified for correctness by the authors before being used in the experiments. We take full responsibility for the final content of this work, including all text, claims, code, and results.

% \subsubsection*{Acknowledgments}
% TODO: funding / compute acknowledgments (hidden in the anonymous submission).

\bibliography{references-revision}

@article{rives2021biological,
  title={Biological structure and function emerge from scaling unsupervised learning to 250 million protein sequences},
  author={Rives, Alexander and Meier, Joshua and Sercu, Tom and Goyal, Siddharth and Lin, Zeming and Liu, Jason and Guo, Demi and Ott, Myle and Zitnick, C Lawrence and Ma, Jerry and Fergus, Rob},
  journal={Proceedings of the National Academy of Sciences},
  volume={118},
  number={15},
  pages={e2016239118},
  year={2021},
  publisher={National Academy of Sciences}
}

@article{lin2023evolutionary,
  title={Evolutionary-scale prediction of atomic-level protein structure with a language model},
  author={Lin, Zeming and Akin, Halil and Rao, Roshan and Hie, Brian and Zhu, Zhongkai and Lu, Wenting and Smetanin, Nikita and Verkuil, Robert and Kabeli, Ori and Shmueli, Yaniv and others},
  journal={Science},
  volume={379},
  number={6637},
  pages={1123--1130},
  year={2023},
  publisher={American Association for the Advancement of Science}
}

@inproceedings{reimers2019sentence,
  title={Sentence-bert: Sentence embeddings using siamese bert-networks},
  author={Reimers, Nils and Gurevych, Iryna},
  booktitle={Proceedings of the 2019 conference on empirical methods in natural language processing and the 9th international joint conference on natural language processing (EMNLP-IJCNLP)},
  pages={3982--3992},
  year={2019}
}

@article{henderson2017efficient,
  title={Efficient natural language response suggestion for smart reply},
  author={Henderson, Matthew and Al-Rfou, Rami and Strope, Brian and Sung, Yun-Hsuan and Luk{\'a}cs, L{\'a}szl{\'o} and Guo, Ruiqi and Kumar, Sanjiv and Miklos, Balint and Kurzweil, Ray},
  journal={arXiv preprint arXiv:1705.00652},
  year={2017}
}

@article{su2022cosent,
  title={Cosent: Consistent sentence embedding via similarity ranking},
  author={Huang, Xiang and Peng, Hao and Zou, Dongcheng and Liu, Zhiwei and Li, Jianxin and Liu, Kay and Wu, Jia and Su, Jianlin and Yu, Philip S},
  journal={IEEE/ACM Transactions on Audio, Speech, and Language Processing},
  volume={32},
  pages={2800--2813},
  year={2024},
  publisher={IEEE}
}

@article{mistry2021pfam,
  title={Pfam: The protein families database in 2021},
  author={Mistry, Jaina and Chuguransky, Sara and Williams, Lowri and Qureshi, Matloob and Salazar, Gustavo A and Sonnhammer, Erik LL and Tosatto, Silvio CE and Paladin, Lisanna and Raj, Shriya and Richardson, Lorna J and others},
  journal={Nucleic acids research},
  volume={49},
  number={D1},
  pages={D412--D419},
  year={2021},
  publisher={Oxford University Press}
}

@article{varadi2022alphafold,
  title={{AlphaFold Protein Structure Database}: massively expanding the structural coverage of protein-sequence space with high-accuracy models},
  author={Varadi, Mihaly and Anyango, Stephen and Deshpande, Mandar and Nair, Sreenath and Natassia, Cindy and Yordanova, Galabina and Yuan, David and Stroe, Oana and Wood, Gemma and Laydon, Agata and others},
  journal={Nucleic Acids Research},
  volume={50},
  number={D1},
  pages={D439--D444},
  year={2022},
  publisher={Oxford University Press}
}

@article{szklarczyk2023string,
  title={The STRING database in 2023: protein--protein association networks and functional enrichment analyses for any sequenced genome of interest},
  author={Szklarczyk, Damian and Kirsch, Rebecca and Koutrouli, Mikaela and Nastou, Katerina and Mehryary, Farrokh and Hachilif, Radja and Gable, Annika L and Fang, Tao and Doncheva, Nadezhda T and Pyysalo, Sampo and others},
  journal={Nucleic acids research},
  volume={51},
  number={D1},
  pages={D638--D646},
  year={2023},
  publisher={Oxford University Press}
}

@article{khurana2018deepsol,
  title={{DeepSol}: a deep learning framework for sequence-based protein solubility prediction},
  author={Khurana, Sameer and Rawi, Reda and Kunji, Khalid and Chuang, Gwo-Yu and Bensmail, Halima and Mall, Raghvendra},
  journal={Bioinformatics},
  volume={34},
  number={15},
  pages={2605--2613},
  year={2018},
  publisher={Oxford University Press}
}

@article{teufel2022signalp,
  title={SignalP 6.0 predicts all five types of signal peptides using protein language models},
  author={Teufel, Felix and Almagro Armenteros, Jos{\'e} Juan and Johansen, Alexander Rosenberg and G{\'\i}slason, Magn{\'u}s Halld{\'o}r and Pihl, Silas Irby and Tsirigos, Konstantinos D and Winther, Ole and Brunak, S{\o}ren and von Heijne, Gunnar and Nielsen, Henrik},
  journal={Nature biotechnology},
  volume={40},
  number={7},
  pages={1023--1025},
  year={2022},
  publisher={Nature Publishing Group US New York}
}

@article{ofer2014profet,
  title={{ProFET}: Feature engineering captures high-level protein functions},
  author={Ofer, Dan and Linial, Michal},
  journal={Bioinformatics},
  volume={31},
  number={21},
  pages={3429--3436},
  year={2015},
  publisher={Oxford University Press}
}

@article{rao2019evaluating,
  title={Evaluating protein transfer learning with {TAPE}},
  author={Rao, Roshan and Bhattacharya, Nicholas and Thomas, Neil and Duan, Yan and Chen, Xi and Canny, John and Abbeel, Pieter and Song, Yun S},
  journal={Advances in Neural Information Processing Systems},
  volume={32},
  year={2019}
}

@inproceedings{dallago2021flip,
  title={FLIP: Benchmark tasks in fitness landscape inference for proteins},
  author={Dallago, Christian and Mou, Jody and Johnston, Kadina E and Wittmann, Bruce J and Bhattacharya, Nicholas and Goldman, Samuel and Madani, Ali and Yang, Kevin K},
  booktitle={Thirty-fifth Conference on Neural Information Processing Systems Datasets and Benchmarks Track (Round 2)},
  year={2021},
  url={https://openreview.net/forum?id=p2dMLEwL8tF}
}

@inproceedings{wolf2020transformers,
  title={Transformers: State-of-the-art natural language processing},
  author={Wolf, Thomas and Debut, Lysandre and Sanh, Victor and Chaumond, Julien and Delangue, Clement and Moi, Anthony and Cistac, Pierric and Rault, Tim and Louf, R{\'e}mi and Funtowicz, Morgan and others},
  booktitle={Proceedings of the 2020 conference on empirical methods in natural language processing: system demonstrations},
  pages={38--45},
  year={2020}
}

@article{heinzinger2022contrastive,
  title={Contrastive learning on protein embeddings enlightens midnight zone},
  author={Heinzinger, Michael and Littmann, Maria and Sillitoe, Ian and Bordin, Nicola and Orengo, Christine and Rost, Burkhard},
  journal={NAR genomics and bioinformatics},
  volume={4},
  number={2},
  pages={lqac043},
  year={2022},
  publisher={Oxford University Press}
}

@inproceedings{redl2023optimizing,
  title={Optimizing protein language models with sentence transformers},
  booktitle={Machine Learning in Structural Biology Workshop, NeurIPS},
  author={Redl, Istvan and Airoldi, Fabio and Bottaro, Sandro and Chung, Albert and Dutton, Oliver and Fisicaro, Carlo and Foerch, Patrik and Henderson, Louie and Hoffmann, Falk and Invernizzi, Michele and others},
  year={2023}
}

@article{liu2024plmsearch,
  title={{PLMSearch}: Protein language model powers accurate and fast sequence search for remote homology},
  author={Liu, Wei and Wang, Ziye and You, Ronghui and Xie, Chenghan and Wei, Hong and Xiong, Yi and Yang, Jianyi and Zhu, Shanfeng},
  journal={Nature Communications},
  volume={15},
  number={1},
  pages={2775},
  year={2024},
  publisher={Nature Publishing Group}
}

@article{hong2024dense,
  title={Fast, sensitive detection of protein homologs using deep dense retrieval},
  author={Hong, Liang and Hu, Zhihang and Sun, Siqi and Tang, Xiangru and Wang, Jiuming and Tan, Qingxiong and Zheng, Liangzhen and Wang, Sheng and Xu, Sheng and King, Irwin and others},
  journal={Nature Biotechnology},
  volume={43},
  number={6},
  pages={983--995},
  year={2025},
  publisher={Nature Publishing Group US New York}
}

@article{lin2025gobeacon,
  title={GOBeacon: An ensemble model for protein function prediction enhanced by contrastive learning},
  author={Lin, Weining and Miller, David and Gu, Zhonghui and Orengo, Christine},
  journal={Protein Science},
  volume={34},
  number={7},
  pages={e70182},
  year={2025},
  publisher={Wiley Online Library}
}

@article{vankempen2024foldseek,
  title={Fast and accurate protein structure search with {Foldseek}},
  author={van Kempen, Michel and Kim, Stephanie S and Tumescheit, Charlotte and Mirdita, Milot and Lee, Jeongjae and Gilchrist, Cameron L M and S{\"o}ding, Johannes and Steinegger, Martin},
  journal={Nature Biotechnology},
  volume={42},
  number={2},
  pages={243--246},
  year={2024},
  publisher={Nature Publishing Group}
}

@article{elnaggar2022prottrans,
  title={{ProtTrans}: Toward understanding the language of life through self-supervised learning},
  author={Elnaggar, Ahmed and Heinzinger, Michael and Dallago, Christian and Rehawi, Ghalia and Wang, Yu and Jones, Llion and Gibbs, Tom and Feher, Tamas and Angerer, Christoph and Steinegger, Martin and Bhowmik, Debsindhu and Rost, Burkhard},
  journal={IEEE Transactions on Pattern Analysis and Machine Intelligence},
  volume={44},
  number={10},
  pages={7112--7127},
  year={2022},
  publisher={IEEE}
}

@inproceedings{reimers2020making,
  title={Making monolingual sentence embeddings multilingual using knowledge distillation},
  author={Reimers, Nils and Gurevych, Iryna},
  booktitle={Proceedings of the 2020 conference on empirical methods in natural language processing (EMNLP)},
  pages={4512--4525},
  year={2020}
}

@article{ofer_language_2021,
  title={The language of proteins: {NLP}, machine learning \& protein sequences},
  author={Ofer, Dan and Brandes, Nadav and Linial, Michal},
  journal={Computational and Structural Biotechnology Journal},
  volume={19},
  pages={1750--1758},
  year={2021},
  publisher={Elsevier}
}

@article{suzek2015uniref,
  title={UniRef clusters: a comprehensive and scalable alternative for improving sequence similarity searches},
  author={Suzek, Baris E and Wang, Yuqi and Huang, Hongzhan and McGarvey, Peter B and Wu, Cathy H and UniProt Consortium, the},
  journal={Bioinformatics},
  volume={31},
  number={6},
  pages={926--932},
  year={2015},
  publisher={Oxford University Press}
}

@article{yang_convolutions_2024,
    title = {Convolutions are competitive with transformers for protein sequence pretraining},
    volume = {15},
    issn = {2405-4712, 2405-4720},
    url = {https://www.cell.com/cell-systems/abstract/S2405-4712(24)00029-2},
    doi = {10.1016/j.cels.2024.01.008},
    language = {English},
    number = {3},
    urldate = {2026-05-03},
    journal = {Cell Systems},
    publisher = {Elsevier},
    author = {Yang, Kevin K. and Fusi, Nicolo and Lu, Alex X.},
    month = mar,
    year = {2024},
    pages = {286--294.e2},
}

@article{heinzinger_bilingual_2024,
  title={Bilingual language model for protein sequence and structure},
  author={Heinzinger, Michael and Weissenow, Konstantin and Sanchez, Joaquin Gomez and Henkel, Adrian and Mirdita, Milot and Steinegger, Martin and Rost, Burkhard},
  journal={NAR Genomics and Bioinformatics},
  volume={6},
  number={4},
  pages={lqae150},
  year={2024},
  publisher={Oxford University Press}
}

@article{barrio-hernandez_clustering_2023,
    title = {Clustering predicted structures at the scale of the known protein universe},
    volume = {622},
    copyright = {2023 The Author(s)},
    issn = {1476-4687},
    url = {https://www.nature.com/articles/s41586-023-06510-w},
    doi = {10.1038/s41586-023-06510-w},
    language = {en},
    number = {7983},
    urldate = {2026-05-04},
    journal = {Nature},
    publisher = {Nature Publishing Group},
    author = {Barrio-Hernandez, Inigo and Yeo, Jingi and Jänes, Jürgen and Mirdita, Milot and Gilchrist, Cameron L. M. and Wein, Tanita and Varadi, Mihaly and Velankar, Sameer and Beltrao, Pedro and Steinegger, Martin},
    month = oct,
    year = {2023},
    pages = {637--645},
}

@article{Jiang2016,
  title={An expanded evaluation of protein function prediction methods shows an improvement in accuracy},
  author={Jiang, Yuxiang and Oron, Tal Ronnen and Clark, Wyatt T and Bankapur, Asma R and D’Andrea, Daniel and Lepore, Rosalba and Funk, Christopher S and Kahanda, Indika and Verspoor, Karin M and Ben-Hur, Asa and others},
  journal={Genome biology},
  volume={17},
  number={1},
  pages={184},
  year={2016},
  publisher={Springer}
}

@article{ofer_protein_2025,
  title={Protein language models expose viral immune mimicry},
  author={Ofer, Dan and Linial, Michal},
  journal={Viruses},
  volume={17},
  number={9},
  pages={1199},
  year={2025},
  publisher={MDPI}
}

@article{michael-pitschaze_detecting_2024,
    title = {Detecting anomalous proteins using deep representations},
    volume = {6},
    issn = {2631-9268},
    url = {https://doi.org/10.1093/nargab/lqae021},
    doi = {10.1093/nargab/lqae021},
    number = {1},
    urldate = {2024-03-02},
    journal = {NAR Genomics and Bioinformatics},
    author = {Michael-Pitschaze, Tomer and Cohen, Niv and Ofer, Dan and Hoshen, Yedid and Linial, Michal},
    month = mar,
    year = {2024},
    pages = {lqae021},
}

@article{Soding2005,
  title={Protein homology detection by HMM--HMM comparison},
  author={S{\"o}ding, Johannes},
  journal={Bioinformatics},
  volume={21},
  number={7},
  pages={951--960},
  year={2005},
  publisher={Oxford University Press}
}

@article{steinegger_mmseqs2_2017,
    title = {{MMseqs2} enables sensitive protein sequence searching for the analysis of massive data sets},
    volume = {35},
    issn = {1087-0156, 1546-1696},
    url = {http://www.nature.com/articles/nbt.3988},
    doi = {10.1038/nbt.3988},
    language = {en},
    number = {11},
    urldate = {2020-11-21},
    journal = {Nature Biotechnology},
    author = {Steinegger, Martin and Söding, Johannes},
    month = nov,
    year = {2017},
    pages = {1026--1028},
}

@inproceedings{ouyangzhang2025ism,
  title={Distilling Structural Representations into Protein Sequence Models},
  author={Ouyang-Zhang, Jeffrey and Gong, Chengyue and Zhao, Yue and Kr{\"a}henb{\"u}hl, Philipp and Klivans, Adam and Diaz, Daniel J},
  booktitle={International Conference on Learning Representations (ICLR)},
  year={2025}
}

@article{zhang2024esms,
  title={Structure-informed protein language model},
  author={Zhang, Zuobai and Lu, Jiarui and Chenthamarakshan, Vijil and Lozano, Aur{\'e}lie and Das, Payel and Tang, Jian},
  journal={arXiv preprint arXiv:2402.05856},
  year={2024}
}

@article{wang2025splm,
  title={{S-PLM}: Structure-Aware Protein Language Model via Contrastive Learning Between Sequence and Structure},
  author={Wang, Duolin and Pourmirzaei, Mahdi and Abbas, Usman L and Zeng, Shuai and Manshour, Negin and Esmaili, Farzaneh and Poudel, Biplab and Jiang, Yuexu and Shao, Qing and Chen, Jin and Xu, Dong},
  journal={Advanced Science},
  volume={12},
  number={5},
  pages={e2404212},
  doi={10.1002/advs.202404212},
  year={2025},
  publisher={Wiley}
}

@inproceedings{calef2026magneton,
  title={Greater than the Sum of Its Parts: Building Substructure into Protein Encoding Models},
  author={Calef, Robert and Liang, Arthur and Kellis, Manolis and Zitnik, Marinka},
  booktitle={International Conference on Learning Representations (ICLR)},
  note={arXiv:2512.18114},
  year={2026}
}

@article{su2025protrek,
  title={A trimodal protein language model enables advanced protein searches},
  author={Su, Jin and He, Yan and You, Shiyang and Jiang, Shiyu and Zhou, Xibin and Zhang, Xuting and Wang, Yuxuan and Su, Xining and Tolstoy, Igor and Chang, Xing and others},
  journal={Nature Biotechnology},
  pages={1--7},
  year={2025},
  publisher={Nature Publishing Group US New York}
}

@inproceedings{khattab2020colbert,
  title={{ColBERT}: Efficient and Effective Passage Search via Contextualized Late Interaction over {BERT}},
  author={Khattab, Omar and Zaharia, Matei},
  booktitle={Proceedings of the 43rd International ACM SIGIR Conference on Research and Development in Information Retrieval},
  pages={39--48},
  year={2020}
}

@article{esmc2024,
  title={Language modeling materializes a world model of protein biology},
  author={Candido, Salvatore and Hayes, Thomas and Derry, Alexander and Rao, Roshan and Lin, Zeming and Verkuil, Robert and Wu, Bryan Z and Lee, Jin Sub and Bruguera, Elise S and Keval, Jehan A and others},
  journal={bioRxiv},
  pages={2026--06},
  year={2026},
  publisher={Cold Spring Harbor Laboratory}
}

@article{notin2023proteingym,
  title={Proteingym: Large-scale benchmarks for protein fitness prediction and design},
  author={Notin, Pascal and Kollasch, Aaron and Ritter, Daniel and Van Niekerk, Lood and Paul, Steffanie and Spinner, Han and Rollins, Nathan and Shaw, Ada and Orenbuch, Rose and Weitzman, Ruben and others},
  journal={Advances in neural information processing systems},
  volume={36},
  pages={64331--64379},
  year={2023}
}

@article{xu2022peer,
  title={Peer: a comprehensive and multi-task benchmark for protein sequence understanding},
  author={Xu, Minghao and Zhang, Zuobai and Lu, Jiarui and Zhu, Zhaocheng and Zhang, Yangtian and Chang, Ma and Liu, Runcheng and Tang, Jian},
  journal={Advances in Neural Information Processing Systems},
  volume={35},
  pages={35156--35173},
  year={2022}
}

@article{bernett2024ppi,
  title={Cracking the black box of deep sequence-based protein--protein interaction prediction},
  author={Bernett, Judith and Blumenthal, David B and List, Markus},
  journal={Briefings in Bioinformatics},
  volume={25},
  number={2},
  pages={bbae076},
  year={2024},
  publisher={Oxford University Press}
}

@article{eddy2011hmmer,
  title={Accelerated profile HMM searches},
  author={Eddy, Sean R},
  journal={PLoS computational biology},
  volume={7},
  number={10},
  pages={e1002195},
  year={2011},
  publisher={Public Library of Science San Francisco, USA}
}

@article{fox2014scope,
  title={{SCOPe}: Structural Classification of Proteins---extended, integrating {SCOP} and {ASTRAL} data and classification of new structures},
  author={Fox, Naomi K and Brenner, Steven E and Chandonia, John-Marc},
  journal={Nucleic Acids Research},
  volume={42},
  number={D1},
  pages={D304--D309},
  year={2014},
  publisher={Oxford University Press}
}

@article{sillitoe2021cath,
  title={{CATH}: increased structural coverage of functional space},
  author={Sillitoe, Ian and Bordin, Nicola and Dawson, Natalie and Waman, Vaishali P and Ashford, Paul and Scholes, Harry M and Pang, Camilla S M and Woodridge, Laurel and Rauer, Clemens and Sen, Neeladri and others},
  journal={Nucleic Acids Research},
  volume={49},
  number={D1},
  pages={D266--D273},
  year={2021},
  publisher={Oxford University Press}
}
\bibliographystyle{iclr2027_conference}

\newpage
% \appendix switches section numbering to A, B, C so cross-references read
% "Appendix C" rather than "Appendix 12".
\appendix
\section*{Appendix}

% Appendix. Regenerated tables in this file are produced from the result
% artifacts by the scripts named in each caption; do not hand-edit their cells.

\section{Model lineage and training details}
\label{app:training}

\subsection{Model lineage}

\begin{table}[h]
  \caption{Every ProtSent arm in this paper. In the main text, ProtSent-V2 is called ProtSent, ProtSent-V2.5 is called ProtSent+DMS, and ProtSent-V1 is the initial configuration. ``dc40'' is the decontaminated corpus of Section~\ref{sec:method:decontam}. ProtSent-V2 arms are the primary models; V2.5 and the ESM-C arm additionally consume a DMS corpus that was \emph{not} put through the MMseqs2 decontamination pass (limitation (vi), Appendix~\ref{app:limitations}).}
  \label{tab:lineage}
  \centering
  \small
  \begin{tabular}{llllc}
    \toprule
    Model & Backbone & Corpus & DMS & Role \\
    \midrule
    ProtSent-V1 35M / 150M   & ESM-2            & unfiltered & yes & initial run \\
    \textbf{ProtSent-V2 35M / 150M} & ESM-2      & dc40       & no  & \textbf{primary} \\
    ProtSent-V2.5 35M / 150M & continues V2      & dc40 + DMS & yes & DMS experiment \\
    ProtSent-V2.5-35M noGOR  & continues V2      & dc40 + DMS & yes & GOR ablation \\
    ProtSent-ESMC-300M       & ESM-C 300M        & dc40 + DMS & yes & second model family \\
    \bottomrule
  \end{tabular}
\end{table}

\subsection{Hyperparameters}

All models are trained with SentenceTransformers~\citep{reimers2019sentence} over HuggingFace Transformers~\citep{wolf2020transformers}.

\begin{table}[h]
  \caption{Training configurations. The V1 columns are the initial runs and are reported for provenance; the 1{,}024 figure there is an \emph{optimizer} batch formed by gradient accumulation, which does not share in-batch negatives, so the MNRL negative pool was the per-device batch (64 and 16). The V2 line uses a cached MNRL implementation that realizes a true 1{,}024-example batch per device.}
  \label{tab:hyperparams}
  \centering
  \small
  \setlength{\tabcolsep}{4pt}
  \begin{tabular}{lcccc}
    \toprule
    & \multicolumn{2}{c}{ProtSent-V1 (initial)} & \multicolumn{2}{c}{ProtSent-V2} \\
    \cmidrule(lr){2-3} \cmidrule(lr){4-5}
    Hyperparameter & 35M & 150M & 35M & 150M \\
    \midrule
    Loss                       & MNRL & MNRL & CachedMNRL & CachedMNRL \\
    Per-device batch           & 64   & 16   & 1024 & 1024 \\
    Gradient accumulation      & 16   & 64   & 1    & 1 \\
    MNRL mini-batch            & --  & --  & 256  & 64 \\
    \textbf{In-batch negatives} & \textbf{64} & \textbf{16} & \textbf{1024} & \textbf{1024} \\
    Learning rate              & $3\times10^{-4}$ & $2\times10^{-4}$ & $2\times10^{-4}$ & $2\times10^{-4}$ \\
    LR scheduler               & \multicolumn{4}{c}{Cosine with min LR} \\
    Warmup steps               & 500 & 1{,}000 & 1{,}000 & 300 \\
    Optimizer                  & \multicolumn{4}{c}{AdamW (fused)}  \\
    Dropout                    & \multicolumn{4}{c}{0.1} \\
    Multi-dataset sampler      & round-robin & round-robin & proportional & proportional \\
    Synthetic hard negatives   & yes & yes & \textbf{no} & \textbf{no} \\
    Pairs per cluster ($k$)    & 8 & 8 & 8 & 5 \\
    Attention                  & sdpa & sdpa & flash-attn-2 & flash-attn-2 \\
    GPUs                       & 1 & 1 & 7 & 6 \\
    Steps                      & -- & -- & -- & 3{,}890 \\
    Wall clock                 & $\sim$4\,h & $\sim$1.3\,d & 10\,h 53\,m & $\sim$53\,h \\
    \bottomrule
  \end{tabular}
\end{table}

\paragraph{ProtSent-V2.5.} A continuation pass from the V2 weights. It changes several settings at once relative to V2, so a V2.5 $-$ V2 difference is not attributable to any one of them: the DMS CoSENT target is added ($\sim$6.5\% of steps at 35M), a Global Orthogonal Regularization term is enabled, the learning rate drops to $5\times10^{-5}$ on a single half-cosine so the run ends annealed rather than at peak, the pairs-per-cluster budget and the shuffle seeds change, and the 35M run is the first to train at a genuine 512-residue cap (see below). The corpus is a fresh draw of the same sources: at 35M, 15.3M pairs of which we measure roughly 28\% to be rows V2 never saw. The GOR-off ablation isolates that one term and finds it contributes nothing measurable (15 ties of 19 tasks; SCOPe-40 R@1 $0.692$ off against $0.690$ on) at an $11.7\%$ step-time cost. This rules out GOR as the source of the V2.5 movement reported in Section~\ref{sec:ablations}; the DMS target, the annealed schedule, the 512-residue cap and the fresh data draw remain confounded with one another.

% CONFIRM: Table~\ref{tab:lineage}, Table~\ref{tab:training-data}, Appendix B.5 and limitation (vi) all say this arm trained with the DMS source; if the run config says otherwise, revert this clause and fix those four places instead.
\paragraph{ProtSent-ESMC-300M.} The V2 recipe, with the DMS source added as in V2.5 (Table~\ref{tab:lineage}), applied to \texttt{Synthyra/ESMplusplus\_small}, a name-for-name port of vanilla ESM-C-300M, the same checkpoint that serves as ISM-C-300M's matched control, with $k{=}10$ pairs per cluster, 7{,}216 steps, GOR and Matryoshka off. No architecture-specific code was needed.

\paragraph{Training-time sequence length.} Every run in Table~\ref{tab:hyperparams} passed \texttt{--max\_seq\_length 512}, and none of them applied it. The flag reached the SentenceTransformers \texttt{Transformer} module, but the custom-code backbone branches then swap that module's tokenizer for the backbone's own, whose \texttt{model\_max\_length} is effectively unbounded; SentenceTransformers reads the truncation limit off the tokenizer, so the requested 512 was discarded and every batch padded to the longest sequence in it. Measured directly, a 512-pair batch arrived as $(512, 1561)$. \textbf{V1, V2-35M and V2-150M therefore trained on untruncated sequences}, and since AFDB and STRING have roughly $29\%$ of sequences longer than 512 residues, those runs saw materially longer inputs than the stated configuration. Evaluation is unaffected, since the benchmark suite pins a 1{,}024-residue cap explicitly, so the consequence is a train/eval length mismatch throughout rather than an invalid measurement. ProtSent-V2.5 is the first run at a genuine 512.

\subsection{Benchmark protocol details}
\label{app:protocol}

\paragraph{Probes.} 3-NN is \texttt{KNeighborsClassifier}/\texttt{KNeighborsRegressor} with $k{=}3$, \textbf{uniform} weighting and Euclidean distance at scikit-learn defaults. The linear probe is standardized logistic regression (one-vs-rest for multiclass) or ridge with $\alpha{=}1$. Encoders are frozen in both cases and never updated.

\paragraph{Pair-input tasks.} For protein--protein interaction, the two partners are embedded independently and their vectors concatenated before the probe. Peptide--HLA binding is \emph{not} a two-input task in this suite: the dataset supplies a single field holding a pipe-joined HLA pseudo-sequence and peptide, so no combination operator is involved.

\paragraph{Probe-invariant rows.} Enzyme commission, GO molecular function and SCOPe-40 retrieval use a built-in evaluator and ignore the requested probe type, so their 3-NN and linear rows are identical by construction. They are a single measurement printed twice.

\paragraph{Tasks excluded from the tallies.} Antibiotic resistance, remote homology and temperature stability have an empty AUC column, because one-vs-rest AUC is undefined when the test split contains a class absent from training. Every tally in this paper covers the other 20 tasks and reports these three separately by accuracy. Substituting an accuracy fallback to reach 23 shifts different arms by different amounts, moving ISM-C's linear median by $-0.006$ and ours by $+0.001$, so we do not use it as a headline.

\paragraph{The remote-homology split.} It is TAPE's three holdouts pooled (718 fold, 1{,}254 superfamily and 1{,}272 family test sequences, 3{,}244 in all) and is \emph{not} hierarchy-disjoint. It is also the set Pfam and AlphaFold~DB were filtered against (Section~\ref{sec:method:decontam}).

\paragraph{Thermostability split.} Thermostability is configured with \texttt{auto\_split}: the suite makes a seeded 80/20 split of \emph{train} and treats the 20\% as the test split. Arms are therefore comparable to each other on that task but not to published FLIP numbers.

\section{Dataset construction details}
% Long \texttt{} resource paths below are set ragged-right: justified text cannot
% break them and leaves large interword gaps.
\newcommand{\srcpar}[1]{{\raggedright #1\par}}
\label{app:datasets}

\begin{table}[h]
  \caption{Training sources after filtering and decontamination. ``Group label'' is the positive-pair criterion for MNRL sources; STRING is pair-native and DMS uses continuous CoSENT scores. Construction details are in Appendix~\ref{app:datasets}. The DMS source is used by ProtSent-V1, ProtSent-V2.5 and ProtSent-ESMC-300M; the primary ProtSent-V2 models train without it.}
  \label{tab:training-data}
  \centering
  \footnotesize
  \setlength{\tabcolsep}{4pt}
  \begin{tabular}{lrll}
    \toprule
    Source & Rows after filtering & Group label & Loss \\
    \midrule
    Pfam (linclust@70\%)        & 27.93M domains & Pfam family ID & MNRL \\
    AFDB50, Foldseek-grouped    & 126.30M sequences & Foldseek structural cluster & MNRL \\
    STRING-DB v12 PPI           & 15.0M pairs (sampled) & Pair-native & MNRL \\
    ProteinGym DMS / clinical   & 2.18M pairs & Continuous score & CoSENT \\
    \midrule
    \textit{Pfam hard negatives} & \textit{1.8M anchors} & \textit{explicit negative} & \textit{MNRL (ablation only)} \\
    \bottomrule
  \end{tabular}
\end{table}

Each training parquet is reproduced by
\texttt{python data\_prep.py --dataset X} at default flags (with one
exception, noted below for AFDB).

\textbf{The ``Final'' counts in each subsection below are
pre-decontamination.} They describe the corpus as constructed; the
decontamination pass of Section~\ref{sec:method:decontam} is applied
afterwards, by a separate script, and reduces them to the figures in
Table~\ref{tab:training-data}: Pfam 28{,}530{,}684~$\rightarrow$~27{,}929{,}772,
AlphaFold~DB 135{,}404{,}259~$\rightarrow$~126{,}301{,}607 and STRING
76{,}070{,}154~$\rightarrow$~71{,}891{,}417 rows, of which a 15M subsample is used
for training. The two sets of numbers differ in scope. The
construction counts are per unique sequence or pair at the point each parquet
is written, whereas the decontamination counts are rows of the training files,
so the two are reported separately rather than reconciled into one column. The DMS
parquet was \textbf{not} put through this pass.

\subsection{Pfam (\texttt{pfam\_sorted.parquet})}
\raggedright \textbf{Source.} \texttt{Pfam-A.fasta.gz} and \texttt{Pfam-A.clans.tsv.gz}
from \texttt{ftp.ebi.ac.uk/pub/databases/Pfam/current\_release}. Headers
parse to \texttt{(domain\_id, family\_id, sequence)} with the PFxxxxx
version stripped.

\textbf{Pipeline.}
(1) MMseqs2 \texttt{easy-linclust --min-seq-id 0.7 --cov-mode 1 -c 0.8}
on the raw Pfam-A FASTA (global asymmetric clustering; family members
assigned to representatives from other families are dropped). (2)
Left-join family $\to$ clan; orphan families inherit
\texttt{clan\_id := family\_id}. (3) Drop singleton families, from which no positive pair can be formed (they enter MNRL only as in-batch negatives, which the other families already supply). (4) Sort by
\texttt{(clan\_id, family\_id)} so windowed slicing preserves clan-level
diversity during training.

\textbf{Final.} 32{,}943{,}498 sequences, 26{,}796 families, 15{,}284
clans (input: $\sim$62M raw Pfam-A FASTA records).

\subsection{Pfam hard negatives (\texttt{pfam\_hard\_negatives.parquet})}
\raggedright \textbf{Source.} \texttt{pfam\_sorted.parquet} above plus
\texttt{Pfam-A.hmm.gz} (parsed with \texttt{pyhmmer}).

\textbf{PSSM.} For each Pfam HMM, take match-state emissions $e_{i,a}$ and
the pyhmmer amino-acid background $b_a$ (both clipped at $10^{-9}$), set
$S_{i,a} = \log_2 e_{i,a} - \log_2 b_a$.

\textbf{Eligibility.} Drop families whose median sequence length differs
from the HMM model length by more than $10\%$. Per row: keep
$5 < L < 1024$; cap at $100$ sequences/family.

\textbf{Sampling.} An \emph{anchor} is a Pfam family member selected from \texttt{pfam\_sorted.parquet} (up to 100 per family) to serve as the wild-type sequence that is mutated. For an anchor of length $L$, use direct map $i \to i$
for $i < \min(L, \text{model\_len})$; compute
$\Delta S_{i,a} = S_{i,a} - S_{i, \text{wt}(i)}$ and zero-out
self-substitutions and non-standard residues. The candidate pool is
positions $(i,a)$ with $\Delta S_{i,a} < -1.0$. Rejection-sample $k$-tuples
(per-family seed $42 + i$, up to $2{,}048$ proposals per $k$) with
$k_{\min} = \max(6, \lceil -16.0 / \min_i \Delta S_i^{(\min)} \rceil)$ and
$k_{\max} \leq 50$, accepting the first proposal that satisfies
$\sum_i \Delta S_i \leq -16.0$ and pairwise position spacing
$\geq \max(6, L/8)$.

\textbf{Final.} 1{,}821{,}068 anchors over 24{,}047 families
(of 26{,}796 in \texttt{pfam\_sorted}, after the $\pm 10\%$ length filter
and the per-family cap of $100$); 1{,}210{,}230 anchors ($66\%$) carry a
non-null \texttt{hard\_negative}.

\textbf{Audit.}
The log-odds budget above is a proxy for ``breaks the family profile'', and we
tested it directly by running HMMER over the generator's own output. The test failed:
\textbf{543 of 543} sampled mutants still matched the family they were meant to
be negatives for, at a worst E-value of $3.8\times10^{-8}$, retaining $88.7\%$
of the wild-type bit score. The sampled mutants were, in effect, slightly noisier
positives. The generator is rewritten to mutate until HMMER no longer
reports the mutant above threshold (2{,}396 of 2{,}396 regenerated mutants now
have $E > 1.0$, at a median identity to the anchor of $85.8\%$), but no model in
this paper is trained on the corrected version. Independently of the audit, the
ablation in Table~\ref{tab:ablation_full} prefers the model \emph{without} this
source, and ProtSent-V2 trains without it.

\subsection{AFDB Foldseek structural pairs (\texttt{afdb\_sorted.parquet})}
\raggedright \textbf{Sources.} (i) \texttt{willdaspit/afdb\_clustered\_seqs}
(HuggingFace), AFDB sequences with their AFDB50 representative IDs, plus
per-row \texttt{plddt} and \texttt{fragment} flags. (ii)
\texttt{1-AFDBClusters-repId\_entryId\_cluFlag\_taxId.tsv.gz} from
\texttt{afdb-cluster.steineggerlab.workers.dev/v6/}, the upstream Foldseek
S-cluster mapping. We retain \texttt{cluFlag} $\in \{1, 2\}$ (singleton
and small-cluster flags are excluded).

\textbf{Pipeline.}
(1) Lazy-scan HF parquets, filter \texttt{plddt}~$>$~70 and
\texttt{fragment}~$=$~0. (2) Inner-join on
\texttt{HF.repId == Steinegger.entry\_id}; carry the Foldseek
representative as \texttt{group\_id} and the AFDB50 representative as
\texttt{afdb50\_cluster\_id}. (3) Pre-shuffle (seed $42$) then sort by
\texttt{afdb50\_cluster\_id}. (4) Run uncapped: we use
\texttt{--limit\_gb 0} (the CLI default of $25$\,GB caps at $\sim$50M
rows).

\textbf{Final.} 133{,}856{,}004 sequences in 815{,}712 Foldseek structural
clusters spanning 1{,}815{,}626 AFDB50 representatives (input: $\sim$214M
AFDB sequences upstream; the pLDDT~$>70$ + non-fragment filter and the
\texttt{cluFlag} $\in \{1,2\}$ inner-join produce the drop).

\textbf{Disambiguation.} ``AFDB'', ``AFDB50'', and ``AFDB Foldseek
clusters'' name three different objects: the full AlphaFold Database;
AFDB sequences clustered at $50\%$ identity (the AFDB50 release); and AFDB
structural clusters from Foldseek over predicted structures (the v6
S-cluster file). The contrastive positive label is the third; the
underlying sequence resource is the second.

\subsection{STRING-DB PPI (\texttt{stringdb\_train.parquet})}
\raggedright \textbf{Sources.} STRING v12.0 \texttt{protein.sequences.v12.0.fa.gz} and
\texttt{protein.physical.links.full.v12.0.txt.gz} from
\texttt{stringdb-downloads.org}.

\textbf{Pipeline.}
(1) Pre-filter the FASTA to proteins that appear in some link with
\texttt{combined\_score}~$\geq 400$ (STRING medium-confidence; reduces
$\sim$59M $\to \sim$25M proteins). (2) Bernett decontamination: load the
\texttt{Synthyra/bernett\_gold\_ppi} test split, write each test sequence
under a \texttt{BERNETT\_*} ID, run a single MMseqs2
\texttt{easy-linclust} at $50\%$ identity / $80\%$ target coverage on the
union, and drop any STRING protein whose linclust cluster contains any
\texttt{BERNETT\_*} member. (3) Two-stage clustering of survivors (all
\texttt{--cov-mode 1}): \texttt{easy-linclust} at $65\%/85\%$ then
cascaded \texttt{cluster} at $50\%/75\%$ on stage-1 reps, with mappings
composed $\text{member} \to \text{rep}_{65} \to \text{rep}_{50}$. (4) Pair
construction: filter links by score, join both endpoints to rep-50
clusters, drop self-cluster edges, canonicalize unordered cluster pairs,
sort by score descending, deduplicate keeping the highest-scoring edge per
cluster pair. (5) Length filter $[10, 1024]$ on both endpoints; final
shuffle (seed $42$); write only \texttt{(seq1, seq2)}.

\textbf{Two decontamination passes.} Step (2) above is the in-pipeline
filter, \texttt{easy-linclust} at $50\%$ with cluster removal. A separate,
stricter pass applied before training ProtSent-V2 does something else,
\texttt{easy-search} at $40\%$ identity and $80\%$ coverage, removing hit query
IDs rather than whole clusters; it removed a further 4{,}178{,}737 pairs and
produced the corpus the V2 models trained on.

\textbf{Final.} 36{,}502{,}692 pairs over 6.73M unique proteins per
endpoint (input: STRING v12.0 has $\sim$59M proteins and $>$20B physical
links; the score $\geq 400$ filter, Bernett decontam, and cluster-pair
deduplication account for the reduction).

\subsection{ProteinGym DMS / clinical (\texttt{dms\_cosent.parquet})}
\raggedright \textbf{Source.} \texttt{OATML-Markslab/ProteinGym\_v1} on HuggingFace,
splits \texttt{DMS\_substitutions}, \texttt{DMS\_indels},
\texttt{clinical\_substitutions}, \texttt{clinical\_indels}.

\textbf{Pipeline.}
(1) Continuous DMS: per-assay z-score of \texttt{DMS\_score} clipped to
$[-3, 3]$ and rescaled to $[0, 1]$. (2) Clinical: Pathogenic $\to 0.0$,
Benign $\to 1.0$. (3) Drop assays with \texttt{DMS\_id} starting
\texttt{GB1\_} or \texttt{GFP\_AEQVI\_} (overlap with our protein-level
benchmarks). (4) Test-fold drop: if a recognized split column
(\texttt{stage}, \texttt{split}, \texttt{set}, \texttt{fold}) carries
\texttt{train}/\texttt{test} labels, drop test rows; otherwise apply the
supervised benchmark's deterministic per-group $80/20$ split
(\texttt{RandomState(42)}, groups of \texttt{DMS\_id} for DMS rows or
\texttt{protein\_id} for clinical rows). Groups with fewer than $10$ rows
are not split (the supervised benchmark also skips these at evaluation,
so this introduces no leak). Pairs that appear in any test row are also
removed globally as a final pass. (5) Final shuffle (seed $42$). Optional
mutant--mutant intra-assay pairing is disabled.

\textbf{Decontamination status: not filtered, and the overlap is measured.}
This parquet predates the corpus-level decontamination pass and was never run
through it. Its only protections are the assay-level drops in steps (3) and (4)
above and an exact-sequence check, which found no overlap. Exact match is not
sequence identity, so we ran the identity check the other three corpora
received: MMseqs2 \texttt{easy-search} with the 2{,}093{,}266 unique corpus
sequences as query and each DMS-derived test set as target, at $40\%$ identity
and $80\%$ coverage of the target.

\begin{table}[h]
\caption{Identity-level overlap between the DMS training corpus (2{,}093{,}266 unique sequences) and the four benchmarks derived from it, at $40\%$ identity and $80\%$ coverage of the target. GB1 is the one benchmark whose assays are dropped by name, and the one with no overlap.}
\label{app:tab:dms_overlap}
\centering
\small
\begin{tabular}{lrrr}
\toprule
Test set & Test sequences & Corpus sequences flagged & \% of corpus \\
\midrule
Fluorescence (TAPE)   & 27{,}217 & 46{,}422 & 2.218 \\
Stability (Biomap)    & 12{,}851 & 11{,}971 & 0.572 \\
beta-lactamase (PEER) &     520 &  6{,}901 & 0.330 \\
Variant Effect (GB1)  &  1{,}745 &  \textbf{0} & \textbf{0.000} \\
\bottomrule
\end{tabular}
\end{table}

The assay-level drop works where it was applied: GB1 is the
one benchmark whose assays step (3) removes by name, and it is the one with zero
overlap. And it does not generalize: step (3) also drops \texttt{GFP\_AEQVI\_},
yet the TAPE fluorescence test set still matches $2.2\%$ of the corpus, so
related assays outside that name pattern carry the overlap through.
Fluorescence, stability and beta-lactamase therefore have \emph{known}
near-duplicate overlap, at a median identity above $0.9$ among hits, on every arm
that trains on this file: ProtSent-V2.5 at both scales and ProtSent-ESMC-300M.
The primary ProtSent-V2 models do not use this source at all, so no ESM-2 result in the
main text depends on it. Reproduce with
\texttt{analysis/dms\_decontam\_check.py}.

\textbf{Final.} 2{,}175{,}734 pairs over 3{,}576 unique wild-type targets
and 2.09M unique mutants (input: $\sim$3M raw ProteinGym v1 rows across
the four splits; the GB1/GFP assay drop and the supervised-test-fold drop
account for the reduction). CoSENT preserves the score-induced ordering
of pairs rather than training to absolute targets.

\subsection{Compute and reproducibility}
End-to-end on a single 48-thread workstation: STRING two-stage clustering
under 6~h, AFDB join $\sim$30~min on NVMe, Pfam hard-negative generation
$\sim$2k anchors/s on 32 threads. All thresholds, identity cutoffs, and
seeds are fixed in \texttt{data\_prep.py}; re-running with default flags
(plus \texttt{--limit\_gb 0} for AFDB) reproduces every parquet
bit-for-bit against a fixed snapshot of the upstream HuggingFace, EBI,
and STRING-DB releases.

\section{Full per-task results}
\label{app:pertask}

Tables~\ref{app:tab:35m_knn}--\ref{app:tab:300m_linear} give every cell behind the win/tie/loss summaries of Table~\ref{tab:probes}. \textbf{There is one measurement per cell and no confidence intervals.} We report them as measurements and draw no inferential claim from the aggregate.

\begin{table}[h]
  \caption{Per-task results, ESM-2 35M line, \textbf{3-NN probe}, test split. $^\dagger$ marks the three tasks with an undefined one-vs-rest AUC, excluded from every win/tie/loss tally and scored by accuracy instead. The SCOPe-40 row is probe-invariant (its own evaluator) and is identical in the linear table.}
  \label{app:tab:35m_knn}
  \centering
  \small
  \begin{tabular}{llccccc}
    \toprule
    Task & Metric & ESM-2 35M & V1 & V2 & V2.5 & V2.5 noGOR \\
    \midrule
    AAV Fitness (FLIP) & Spr. & 0.4667 & 0.5553 & 0.5154 & 0.4945 & 0.4821 \\
    Antibiotic Resistance$^\dagger$ & Acc. & 0.9673 & 0.9583 & 0.9650 & 0.9635 & 0.9628 \\
    Binary Subcel. Loc. & AUC & 0.8810 & 0.9089 & 0.8884 & 0.8966 & 0.8954 \\
    Cloning Classification & Spr. & 0.3906 & 0.3774 & 0.3791 & 0.3928 & 0.3912 \\
    EC Classification & F1$_\mathrm{M}$ & 0.5984 & 0.5617 & 0.5924 & 0.5956 & 0.5962 \\
    Enzyme Cat. Efficiency & Spr. & 0.6915 & 0.6826 & 0.6687 & 0.6745 & 0.6749 \\
    Fluorescence (TAPE) & Spr. & 0.3736 & 0.4510 & 0.4568 & 0.4669 & 0.4620 \\
    Material Production & AUC & 0.7684 & 0.7716 & 0.7653 & 0.7662 & 0.7641 \\
    Metal Ion Binding & AUC & 0.7957 & 0.8015 & 0.8158 & 0.8000 & 0.8014 \\
    GO Molecular Function & F1$_\mathrm{M}$ & 0.4590 & 0.4434 & 0.4547 & 0.4506 & 0.4532 \\
    Neuropeptide (NeuroPID) & AUC & 0.9579 & 0.9584 & 0.9601 & 0.9549 & 0.9564 \\
    Optimal pH & Spr. & 0.5821 & 0.5742 & 0.5756 & 0.5655 & 0.5613 \\
    Peptide-HLA Binding & AUC & 0.7496 & 0.7860 & 0.8022 & 0.8033 & 0.8029 \\
    Remote Homology (Fold)$^\dagger$ & Acc. & 0.5835 & 0.6587 & 0.6668 & 0.6640 & 0.6646 \\
    SCOPe-40 retrieval (MAP) & MAP & 0.4210 & 0.5509 & 0.6459 & 0.6521 & 0.6528 \\
    Signal Peptide & AUC & 0.9780 & 0.9777 & 0.9840 & 0.9829 & 0.9815 \\
    Solubility (DeepSol) & AUC & 0.5321 & 0.5413 & 0.5426 & 0.5413 & 0.5454 \\
    Stability (Biomap) & Spr. & 0.6435 & 0.5638 & 0.5961 & 0.5989 & 0.6090 \\
    Subcellular Localization & AUC & 0.8015 & 0.8246 & 0.8294 & 0.8173 & 0.8212 \\
    Temperature Stability$^\dagger$ & Acc. & 0.8613 & 0.8375 & 0.8514 & 0.8514 & 0.8518 \\
    Thermostability (FLIP) & Spr. & 0.4449 & 0.4754 & 0.4367 & 0.4675 & 0.4516 \\
    Variant Effect (GB1) & Spr. & 0.6582 & 0.7107 & 0.7806 & 0.7796 & 0.7910 \\
    Beta-lactamase (PEER) & Spr. & 0.7272 & 0.7676 & 0.7153 & 0.7305 & 0.7261 \\
    \bottomrule
  \end{tabular}
\end{table}

\begin{table}[h]
  \caption{Per-task results, ESM-2 35M line, \textbf{linear probe} (logistic / ridge on standardized frozen features), test split. Conventions as in Table~\ref{app:tab:35m_knn}.}
  \label{app:tab:35m_lin}
  \centering
  \small
  \begin{tabular}{llccccc}
    \toprule
    Task & Metric & ESM-2 35M & V1 & V2 & V2.5 & V2.5 noGOR \\
    \midrule
    AAV Fitness (FLIP) & Spr. & 0.5639 & 0.4362 & 0.2471 & 0.3204 & 0.2998 \\
    Antibiotic Resistance$^\dagger$ & Acc. & 0.9755 & 0.9650 & 0.9665 & 0.9695 & 0.9673 \\
    Binary Subcel. Loc. & AUC & 0.9572 & 0.9404 & 0.9093 & 0.8965 & 0.8932 \\
    Cloning Classification & Spr. & 0.4780 & 0.4420 & 0.4257 & 0.4338 & 0.4334 \\
    EC Classification & F1$_\mathrm{M}$ & 0.5984 & 0.5617 & 0.5924 & 0.5956 & 0.5962 \\
    Enzyme Cat. Efficiency & Spr. & 0.5456 & 0.5122 & 0.5002 & 0.5017 & 0.5075 \\
    Fluorescence (TAPE) & Spr. & 0.5913 & 0.5912 & 0.5883 & 0.6031 & 0.5955 \\
    Material Production & AUC & 0.8399 & 0.8244 & 0.8225 & 0.8223 & 0.8228 \\
    Metal Ion Binding & AUC & 0.7903 & 0.7599 & 0.7466 & 0.7408 & 0.7341 \\
    GO Molecular Function & F1$_\mathrm{M}$ & 0.4590 & 0.4434 & 0.4547 & 0.4506 & 0.4532 \\
    Neuropeptide (NeuroPID) & AUC & 0.9748 & 0.9763 & 0.9594 & 0.9625 & 0.9637 \\
    Optimal pH & Spr. & 0.4864 & 0.4740 & 0.4845 & 0.4598 & 0.4598 \\
    Peptide-HLA Binding & AUC & 0.8628 & 0.8187 & 0.8579 & 0.8590 & 0.8578 \\
    Remote Homology (Fold)$^\dagger$ & Acc. & 0.6868 & 0.6899 & 0.7016 & 0.7047 & 0.7056 \\
    SCOPe-40 retrieval (MAP) & MAP & 0.4210 & 0.5509 & 0.6459 & 0.6521 & 0.6528 \\
    Signal Peptide & AUC & 0.9942 & 0.9949 & 0.9957 & 0.9948 & 0.9952 \\
    Solubility (DeepSol) & AUC & 0.6963 & 0.6929 & 0.6976 & 0.6901 & 0.6930 \\
    Stability (Biomap) & Spr. & 0.4395 & 0.5110 & 0.3878 & 0.4824 & 0.4941 \\
    Subcellular Localization & AUC & 0.9116 & 0.9057 & 0.8921 & 0.8903 & 0.8892 \\
    Temperature Stability$^\dagger$ & Acc. & 0.8973 & 0.8779 & 0.8963 & 0.8970 & 0.8974 \\
    Thermostability (FLIP) & Spr. & 0.5413 & 0.5660 & 0.5564 & 0.5459 & 0.5515 \\
    Variant Effect (GB1) & Spr. & 0.8163 & 0.8247 & 0.8126 & 0.8254 & 0.8266 \\
    Beta-lactamase (PEER) & Spr. & 0.6639 & 0.5762 & 0.6086 & 0.6195 & 0.5926 \\
    \bottomrule
  \end{tabular}
\end{table}

\begin{table}[h]
  \caption{Per-task results, ESM-2 150M line, \textbf{3-NN probe}, test split. Conventions as in Table~\ref{app:tab:35m_knn}.}
  \label{app:tab:150m_knn}
  \centering
  \small
  \begin{tabular}{llcccc}
    \toprule
    Task & Metric & ESM-2 150M & V1 & V2 & V2.5 \\
    \midrule
    AAV Fitness (FLIP) & Spr. & 0.4345 & 0.4663 & 0.5031 & 0.5062 \\
    Antibiotic Resistance$^\dagger$ & Acc. & 0.9680 & 0.9628 & 0.9658 & 0.9688 \\
    Binary Subcel. Loc. & AUC & 0.8964 & 0.9010 & 0.9032 & 0.9027 \\
    Cloning Classification & Spr. & 0.3754 & 0.3789 & 0.3627 & 0.3857 \\
    EC Classification & F1$_\mathrm{M}$ & 0.6466 & 0.6040 & 0.6253 & 0.6205 \\
    Enzyme Cat. Efficiency & Spr. & 0.7060 & 0.6777 & 0.6909 & 0.6857 \\
    Fluorescence (TAPE) & Spr. & 0.3802 & 0.4619 & 0.3839 & 0.4521 \\
    Material Production & AUC & 0.7582 & 0.7666 & 0.7644 & 0.7721 \\
    Metal Ion Binding & AUC & 0.7973 & 0.8147 & 0.8166 & 0.8054 \\
    GO Molecular Function & F1$_\mathrm{M}$ & 0.5116 & 0.4961 & 0.5148 & 0.5011 \\
    Neuropeptide (NeuroPID) & AUC & 0.9777 & 0.9614 & 0.9713 & 0.9728 \\
    Optimal pH & Spr. & 0.5622 & 0.5836 & 0.5892 & 0.5869 \\
    Peptide-HLA Binding & AUC & 0.7565 & 0.7790 & 0.7999 & 0.8008 \\
    Remote Homology (Fold)$^\dagger$ & Acc. & 0.5194 & 0.7047 & 0.6612 & 0.6810 \\
    SCOPe-40 retrieval (MAP) & MAP & 0.4236 & 0.6431 & 0.7046 & 0.7227 \\
    Signal Peptide & AUC & 0.9804 & 0.9740 & 0.9827 & 0.9806 \\
    Solubility (DeepSol) & AUC & 0.5465 & 0.5485 & 0.5263 & 0.5319 \\
    Stability (Biomap) & Spr. & 0.6495 & 0.6097 & 0.6239 & 0.6375 \\
    Subcellular Localization & AUC & 0.8188 & 0.8265 & 0.8358 & 0.8370 \\
    Temperature Stability$^\dagger$ & Acc. & 0.8911 & 0.8480 & 0.8619 & 0.8617 \\
    Thermostability (FLIP) & Spr. & 0.4377 & 0.4606 & 0.4623 & 0.4952 \\
    Variant Effect (GB1) & Spr. & 0.6607 & 0.7983 & 0.7700 & 0.7953 \\
    Beta-lactamase (PEER) & Spr. & 0.7728 & 0.7423 & 0.7619 & 0.7833 \\
    \bottomrule
  \end{tabular}
\end{table}

\begin{table}[h]
  \caption{Per-task results, ESM-2 150M line, \textbf{linear probe}, test split. Conventions as in Table~\ref{app:tab:35m_knn}.}
  \label{app:tab:150m_lin}
  \centering
  \small
  \begin{tabular}{llcccc}
    \toprule
    Task & Metric & ESM-2 150M & V1 & V2 & V2.5 \\
    \midrule
    AAV Fitness (FLIP) & Spr. & 0.5887 & 0.3976 & 0.4509 & 0.3646 \\
    Antibiotic Resistance$^\dagger$ & Acc. & 0.9814 & 0.9747 & 0.9725 & 0.9740 \\
    Binary Subcel. Loc. & AUC & 0.9499 & 0.9351 & 0.9234 & 0.9175 \\
    Cloning Classification & Spr. & 0.5071 & 0.4783 & 0.4345 & 0.4406 \\
    EC Classification & F1$_\mathrm{M}$ & 0.6466 & 0.6040 & 0.6253 & 0.6205 \\
    Enzyme Cat. Efficiency & Spr. & 0.5466 & 0.5478 & 0.5432 & 0.5403 \\
    Fluorescence (TAPE) & Spr. & 0.5799 & 0.6079 & 0.5908 & 0.6059 \\
    Material Production & AUC & 0.8528 & 0.8400 & 0.8344 & 0.8343 \\
    Metal Ion Binding & AUC & 0.7920 & 0.7891 & 0.7690 & 0.7610 \\
    GO Molecular Function & F1$_\mathrm{M}$ & 0.5116 & 0.4961 & 0.5148 & 0.5011 \\
    Neuropeptide (NeuroPID) & AUC & 0.9802 & 0.9762 & 0.9714 & 0.9724 \\
    Optimal pH & Spr. & 0.5315 & 0.5170 & 0.5034 & 0.4989 \\
    Peptide-HLA Binding & AUC & 0.8630 & 0.8374 & 0.8712 & 0.8747 \\
    Remote Homology (Fold)$^\dagger$ & Acc. & 0.7500 & 0.7401 & 0.7503 & 0.7503 \\
    SCOPe-40 retrieval (MAP) & MAP & 0.4236 & 0.6431 & 0.7046 & 0.7227 \\
    Signal Peptide & AUC & 0.9942 & 0.9941 & 0.9961 & 0.9958 \\
    Solubility (DeepSol) & AUC & 0.7211 & 0.7085 & 0.7076 & 0.7049 \\
    Stability (Biomap) & Spr. & 0.7060 & 0.6987 & 0.6625 & 0.6391 \\
    Subcellular Localization & AUC & 0.9203 & 0.9156 & 0.9079 & 0.9126 \\
    Temperature Stability$^\dagger$ & Acc. & 0.9264 & 0.8908 & 0.9097 & 0.9145 \\
    Thermostability (FLIP) & Spr. & 0.5605 & 0.5581 & 0.5606 & 0.5605 \\
    Variant Effect (GB1) & Spr. & 0.8047 & 0.8448 & 0.8294 & 0.8268 \\
    Beta-lactamase (PEER) & Spr. & 0.7788 & 0.6807 & 0.6454 & 0.6939 \\
    \bottomrule
  \end{tabular}
\end{table}

\begin{table}[h]
  \caption{Per-task results, \textbf{ESM-C 300M line}, \textbf{3-NN probe}, test split. ISM-C-300M is structure-distilled from ESM-C-300M and ProtSent-ESMC-300M applies the ProtSent-V2 recipe plus the DMS source (Table~\ref{tab:lineage}) to the same backbone, so all three rows share an architecture, a parameter count and a tokenizer. Its fluorescence, stability and beta-lactamase cells overlap the DMS corpus (Table~\ref{app:tab:dms_overlap}). $^\dagger$ as in Table~\ref{app:tab:35m_knn}.}
  \label{app:tab:300m_knn}
  \centering
  \small
  \begin{tabular}{llccc}
    \toprule
    Task & Metric & ESM-C 300M & ISM-C 300M & ProtSent-ESMC-300M \\
    \midrule
    AAV Fitness (FLIP) & Spr. & 0.4921 & 0.4835 & 0.4753 \\
    Antibiotic Resistance$^\dagger$ & Acc. & 0.9628 & 0.9420 & 0.9747 \\
    Binary Subcel. Loc. & AUC & 0.8813 & 0.8660 & 0.8873 \\
    Cloning Classification & Spr. & 0.3272 & 0.5071 & 0.3695 \\
    EC Classification & F1$_\mathrm{M}$ & 0.6404 & 0.5274 & 0.6470 \\
    Enzyme Cat. Efficiency & Spr. & 0.6957 & 0.6908 & 0.6827 \\
    Fluorescence (TAPE) & Spr. & 0.3706 & 0.5436 & 0.4488 \\
    Material Production & AUC & 0.7266 & 0.8087 & 0.7684 \\
    Metal Ion Binding & AUC & 0.7369 & 0.6947 & 0.8295 \\
    GO Molecular Function & F1$_\mathrm{M}$ & 0.5164 & 0.4378 & 0.5285 \\
    Neuropeptide (NeuroPID) & AUC & 0.9781 & 0.9673 & 0.9699 \\
    Optimal pH & Spr. & 0.5113 & 0.4525 & 0.6132 \\
    Peptide-HLA Binding & AUC & 0.7220 & 0.7457 & 0.7947 \\
    Remote Homology (Fold)$^\dagger$ & Acc. & 0.3545 & 0.4026 & 0.6810 \\
    SCOPe-40 retrieval (MAP) & MAP & 0.2212 & 0.2733 & 0.7692 \\
    Signal Peptide & AUC & 0.9749 & 0.9679 & 0.9820 \\
    Solubility (DeepSol) & AUC & 0.5944 & 0.6852 & 0.5314 \\
    Stability (Biomap) & Spr. & 0.6935 & 0.5794 & 0.5128 \\
    Subcellular Localization & AUC & 0.7862 & 0.7756 & 0.8380 \\
    Temperature Stability$^\dagger$ & Acc. & 0.9257 & 0.8394 & 0.8574 \\
    Thermostability (FLIP) & Spr. & 0.4497 & 0.4447 & 0.4641 \\
    Variant Effect (GB1) & Spr. & 0.7989 & 0.7199 & 0.8149 \\
    Beta-lactamase (PEER) & Spr. & 0.8239 & 0.8351 & 0.7697 \\
    \bottomrule
  \end{tabular}
\end{table}

\begin{table}[h]
  \caption{Per-task results, \textbf{ESM-C 300M line}, \textbf{linear probe}, test split. ISM-C-300M is structure-distilled from ESM-C-300M and ProtSent-ESMC-300M applies the ProtSent-V2 recipe plus the DMS source (Table~\ref{tab:lineage}) to the same backbone, so all three rows share an architecture, a parameter count and a tokenizer. Its fluorescence, stability and beta-lactamase cells overlap the DMS corpus (Table~\ref{app:tab:dms_overlap}). $^\dagger$ as in Table~\ref{app:tab:35m_knn}.}
  \label{app:tab:300m_linear}
  \centering
  \small
  \begin{tabular}{llccc}
    \toprule
    Task & Metric & ESM-C 300M & ISM-C 300M & ProtSent-ESMC-300M \\
    \midrule
    AAV Fitness (FLIP) & Spr. & 0.6058 & 0.5814 & 0.4498 \\
    Antibiotic Resistance$^\dagger$ & Acc. & 0.9814 & 0.9658 & 0.9814 \\
    Binary Subcel. Loc. & AUC & 0.9519 & 0.9394 & 0.9058 \\
    Cloning Classification & Spr. & 0.5129 & 0.5736 & 0.4463 \\
    EC Classification & F1$_\mathrm{M}$ & 0.6404 & 0.5274 & 0.6470 \\
    Enzyme Cat. Efficiency & Spr. & 0.6073 & 0.5845 & 0.5947 \\
    Fluorescence (TAPE) & Spr. & 0.6155 & 0.6585 & 0.6393 \\
    Material Production & AUC & 0.8454 & 0.8771 & 0.8338 \\
    Metal Ion Binding & AUC & 0.7882 & 0.7662 & 0.7608 \\
    GO Molecular Function & F1$_\mathrm{M}$ & 0.5164 & 0.4378 & 0.5285 \\
    Neuropeptide (NeuroPID) & AUC & 0.9915 & 0.9837 & 0.9725 \\
    Optimal pH & Spr. & 0.5238 & 0.5079 & 0.5246 \\
    Peptide-HLA Binding & AUC & 0.8659 & 0.8946 & 0.8768 \\
    Remote Homology (Fold)$^\dagger$ & Acc. & 0.6995 & 0.6705 & 0.7534 \\
    SCOPe-40 retrieval (MAP) & MAP & 0.2212 & 0.2733 & 0.7692 \\
    Signal Peptide & AUC & 0.9942 & 0.9957 & 0.9935 \\
    Solubility (DeepSol) & AUC & 0.7538 & 0.8222 & 0.6981 \\
    Stability (Biomap) & Spr. & 0.7550 & 0.7003 & 0.4591 \\
    Subcellular Localization & AUC & 0.9231 & 0.9185 & 0.8978 \\
    Temperature Stability$^\dagger$ & Acc. & 0.9562 & 0.9333 & 0.9198 \\
    Thermostability (FLIP) & Spr. & 0.5498 & 0.5517 & 0.5366 \\
    Variant Effect (GB1) & Spr. & 0.8165 & 0.8412 & 0.8396 \\
    Beta-lactamase (PEER) & Spr. & 0.8324 & 0.8082 & 0.6816 \\
    \bottomrule
  \end{tabular}
\end{table}

\section{Alignment baselines, per task}
\label{app:alignment}

Table~\ref{app:tab:alignment} is the evidence for the trade-off argument in Section~\ref{sec:discussion}. Read the coverage columns alongside the scores: where annotation transfers by homology and coverage is high, alignment beats every embedding arm we scored, and where coverage collapses the reported score stops being a measurement of the method.

\begin{table}[h]
  \caption{Alignment baselines across the benchmark: HMMER phmmer (\texttt{-E 10}, default filters, rank by bitscore) and MMseqs2 (\texttt{-s 7.5 -e 10 --max-seqs 300}), scored through the same code as the embeddings with no-hit queries counted as failures. \emph{Hit coverage} is the fraction of test queries for which the tool returned any hit at all; where it is low, the reported score is largely a property of the fallback rather than of the method. RhlA (6-residue mutation-site strings) is the extreme case and is not part of the 23-task embedding suite. Antibiotic resistance, remote homology and temperature stability are scored by AUC here but by accuracy in Tables~\ref{app:tab:35m_knn}--\ref{app:tab:300m_linear}, so those three rows are not comparable across the two.}
  \label{app:tab:alignment}
  \centering
  \small
  \begin{tabular}{llcccc}
    \toprule
    & & \multicolumn{2}{c}{HMMER phmmer} & \multicolumn{2}{c}{MMseqs2} \\
    \cmidrule(lr){3-4} \cmidrule(lr){5-6}
    Task & Metric & Score & Coverage & Score & Coverage \\
    \midrule
    AAV Fitness (FLIP) & Spr. & 0.370 & 1.000 & 0.402 & 1.000 \\
    Antibiotic Resistance & AUC & 0.955 & 0.989 & 0.954 & 0.995 \\
    Beta-lactamase (PEER) & Spr. & 0.797 & 1.000 & 0.803 & 1.000 \\
    Binary Subcel. Loc. & AUC & 0.670 & 0.622 & 0.683 & 0.895 \\
    Cloning Classification & Spr. & 0.190 & 0.927 & 0.171 & 0.967 \\
    EC Classification & F1$_\mathrm{M}$ & 0.723 & 0.945 & 0.710 & 0.990 \\
    Enzyme Cat. Efficiency & Spr. & 0.646 & 0.991 & 0.632 & 0.994 \\
    Fluorescence (TAPE) & Spr. & 0.284 & 1.000 & 0.386 & 1.000 \\
    GO Molecular Function & F1$_\mathrm{M}$ & 0.605 & 0.901 & 0.585 & 0.956 \\
    Material Production & AUC & 0.584 & 0.927 & 0.580 & 0.964 \\
    Metal Ion Binding & AUC & 0.728 & 0.832 & 0.724 & 0.954 \\
    Neuropeptide (NeuroPID) & AUC & 0.893 & 0.828 & 0.901 & 0.955 \\
    Optimal pH & Spr. & 0.554 & 0.937 & 0.546 & 0.987 \\
    Peptide-HLA Binding & AUC & 0.631 & 1.000 & 0.637 & 1.000 \\
    Remote Homology (Fold) & AUC & 0.644 & 0.524 & 0.652 & 0.889 \\
    RhlA Enzyme Mutations & Spr. & -0.089 & 0.004 & -- & 0.000 \\
    SCOPe-40 retrieval & R@10 & 0.596 & 0.687 & 0.564 & 0.882 \\
    Signal Peptide & AUC & 0.811 & 0.675 & 0.796 & 0.770 \\
    Solubility (DeepSol) & AUC & 0.415 & 0.951 & 0.418 & 0.951 \\
    Stability (Biomap) & Spr. & 0.597 & 1.000 & 0.582 & 1.000 \\
    Subcellular Localization & AUC & 0.672 & 0.680 & 0.683 & 0.897 \\
    Temperature Stability & AUC & 0.688 & 0.996 & 0.685 & 0.972 \\
    Thermostability (FLIP) & Spr. & 0.470 & 0.999 & 0.480 & 0.993 \\
    Variant Effect (GB1) & Spr. & 0.717 & 1.000 & 0.717 & 1.000 \\
    \bottomrule
  \end{tabular}
\end{table}

\section{Structural benchmarks: protocol and controls}
\label{app:structural}

\subsection{SCOPe-40}

The corpus is \texttt{tattabio/scope40\_test}: 2{,}207 domains, evaluated all-versus-all with the query excluded. A retrieval is correct when the retrieved domain shares the query's \textbf{family}. Of the 2{,}207 queries, 1{,}693 have at least one non-self same-family protein in the gallery and are the only ones that can succeed at any $k$; all reported figures are means over those. Unrestricted values are exactly the eligible values scaled by $1693/2207$. MAP is computed over the full ranking. No-hit alignment queries are left in an arbitrary tail so they score as failures. The paired intervals for the ESM-C line were computed from a re-encode of the gallery on a different device, using the same per-query scoring and bootstrap code as the ESM-2 comparisons; it reproduces the printed rows to within $0.002$ (ESM-C $0.3715$/$0.5783$/$0.2206$ against $0.3709$/$0.5794$/$0.2212$; ProtSent-ESMC $0.7956$/$0.9545$/$0.7689$ against $0.7974$/$0.9539$/$0.7692$).

\paragraph{Alignment configuration.} We report HMMER phmmer in its filters-off, maximally sensitive configuration (eligible R@1 $0.7525$, R@10 $0.8978$, MAP $0.6067$). A default-filter run is weaker ($0.6970$ / $0.7809$ / $0.4749$), and quoting the weaker one would support the claim that ProtSent leads profile search at top-1, which does not hold at either ESM-2 scale. The identity-stratified analysis in Table~\ref{tab:clean} predates this and uses the default-filter configuration; it is labeled accordingly and is the only place where that configuration enters a comparison. For MMseqs2, sensitivity must always be stated: the default \texttt{-s 5.7} gives SCOPe-40 R@1 $0.385$ against $0.656$ at \texttt{-s 7.5}.

\begin{figure}[h]
  \centering
  \includegraphics[width=0.9\textwidth]{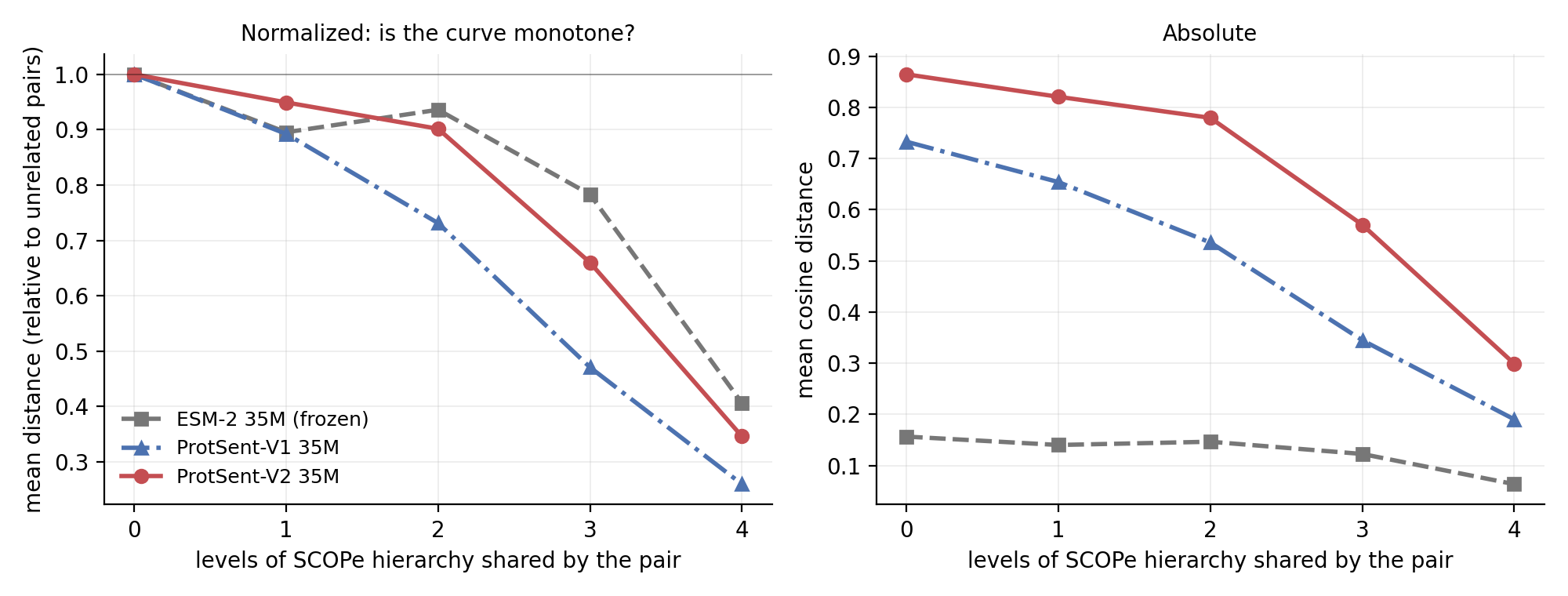}
  \caption{The distance profiles of Table~\ref{tab:geometry}: mean embedding distance against the number of SCOPe hierarchy levels a pair of domains shares, over all pairs among the 2{,}207 domains. \textbf{Left}: each curve divided by its value at zero shared levels, to compare shapes; a representation that encodes the hierarchy should decrease monotonically. ProtSent-V1 and V2 do; the frozen backbone rises between one and two shared levels. \textbf{Right}: unnormalized. The untuned space separates related from unrelated pairs over about $0.09$ cosine, against $0.57$ for ProtSent-V2.}
  \label{fig:hierarchy}
\end{figure}

\paragraph{Initial configuration (V1).} On SCOPe-40 the initial configuration scores R@1/R@10/MAP of $0.585$/$0.851$/$0.551$ at 35M and $0.662$/$0.894$/$0.643$ at 150M, between the frozen base and ProtSent-V2. Filtering the corpora therefore did not reduce retrieval performance: V2 leads V1 by $+0.099\,[+0.076,+0.121]$ and $+0.081\,[+0.060,+0.102]$ Recall@1 at 35M and 150M, although the two differ in several settings at once (Table~\ref{tab:hyperparams}), so this is not a controlled measurement of the filter alone. Clustering statistics for V1 at 35M are ARI $0.417$, NMI $0.900$ and silhouette $-0.044$, and its distance profile is monotone (Figure~\ref{fig:hierarchy}). On remote homology, the task the corpora were filtered against, V1 scores $0.659$ at 35M and $0.705$ at 150M under 3-NN against $0.667$ and $0.661$ for V2, and $0.690$ and $0.740$ under a linear probe against $0.702$ and $0.750$; the 150M 3-NN drop is consistent with the larger unfiltered model having benefited from leakage.

\paragraph{Precision.} At $k{=}10$ over the eligible queries, precision is $0.279$ for ESM-2 35M, $0.329$ for ProtSent-V1 and $0.367$ for ProtSent-V2, against an attainable ceiling of $0.516$ set by the number of same-family domains available. The corresponding alignment figures are $0.245$ for phmmer and $0.207$ for MMseqs2, but this analysis was run against the \emph{default-filter} phmmer configuration, so those two values are not comparable to the filters-off rows of Table~\ref{tab:scope} and we do not use them in the main text.

\begin{table}[h]
  \caption{SCOPe-40 restricted to queries below a maximum identity to our pretraining corpus. Paired bootstrap, ProtSent-V2 35M minus HMMER phmmer. \emph{Note:} this analysis was run against the default-filter phmmer configuration (all-query R@1 $0.697$), not the filters-off configuration quoted in Table~\ref{tab:scope}.}
  \label{tab:clean}
  \centering
  \footnotesize
  \setlength{\tabcolsep}{2.5pt}
  \begin{tabular}{lccc}
    \toprule
    Eligible queries kept & Recall@1 & Recall@10 & MAP \\
    \midrule
    identity $< 0.4$ ($n{=}164$)   & $-0.043\,[-0.128,+0.043]$ & $+0.116\,[+0.049,+0.189]$ & $+0.140\,[+0.075,+0.207]$ \\
    identity $< 0.7$ ($n{=}479$)   & $-0.027\,[-0.073,+0.017]$ & $+0.127\,[+0.090,+0.165]$ & $+0.154\,[+0.117,+0.190]$ \\
    all ($n{=}1{,}693$)            & $-0.012\,[-0.037,+0.012]$ & $+0.141\,[+0.120,+0.162]$ & $+0.171\,[+0.151,+0.191]$ \\
    \bottomrule
  \end{tabular}
\end{table}

\paragraph{Identity stratification and the headroom control.} At 150M, the retrieval gain over the frozen base is if anything \emph{larger} on the queries furthest from our corpus: $\Delta$MAP is $+0.289$ on the 164 queries at identity $[0.2,0.4)$, $+0.303$ on the 315 at $[0.4,0.7)$ and $+0.274$ on the 1{,}214 at $[0.7,1.0]$. The raw per-query Spearman between identity and gain is negative (R@10 $-0.062$, $p{=}0.011$; MAP $-0.083$, $p{=}6\times10^{-4}$), and after controlling for baseline headroom it collapses to a null at 150M ($-0.002$ and $-0.037$) while remaining significantly negative at 35M. Memorization predicts a positive correlation at either scale.

\paragraph{Checkpoint control.} A near-trough checkpoint of each V2 run scores within $0.001$--$0.008$ of the final checkpoint at 35M and $0.002$--$0.021$ at 150M on the SCOPe-40 metrics (the largest gap is MAP at 150M, $0.684$ against $0.705$), so the reported numbers are not an artifact of a schedule that ends at peak learning rate. It is also the reason we treat no structural difference below $0.01$ at 35M or $0.02$ at 150M as resolved.

\subsection{CATH midnight zone: controls}

\begin{table}[h]
  \caption{CATH controls at the Homologous-superfamily level. \emph{L2}: L2-normalizing the vanilla embeddings, which gives them direction-only geometry without training. \emph{Scale}: ESM-2 is flat and non-monotone in parameter count on this task. \emph{Floor}: 3-mer frequency vectors, confirming no composition shortcut. The two ESM-2 35M load paths give bit-identical results, ruling out a code-path artifact.}
  \label{app:tab:cath_controls}
  \centering
  \small
  \begin{tabular}{lcc}
    \toprule
    Arm & raw & L2-normalized \\
    \midrule
    ESM-2 35M                     & 40.7 & 43.3 \\
    ESM-2 35M (plain HF load path) & 40.7 & -- \\
    ESM-2 150M                    & 43.3 & 44.7 \\
    ESM-2 650M                    & 42.7 & -- \\
    ProtSent-V2 35M               & 56.7 & 56.7 \\
    ProtSent-V2 150M              & 62.7 & 62.0 \\
    \midrule
    3-mer frequency floor         & \multicolumn{2}{c}{0.0} \\
    MMseqs2 (published: 35)       & \multicolumn{2}{c}{34.7} \\
    phmmer (sequence, not profile)& \multicolumn{2}{c}{38.0} \\
    \bottomrule
  \end{tabular}
\end{table}

Normalization is worth $+2.7$ and $+1.3$ points to the two vanilla bases and nothing at all to ProtSent, whose geometry is already direction-dominated; scoring vanilla at its best still leaves gaps of $+13.3$ (35M) and $+18.0$ (150M). Neither model has a \texttt{Normalize} module (both are Transformer plus mean pooling), so the pipelines are identical. The published HMMER row in Table~\ref{tab:cath} uses CATH-Gene3D \emph{profiles} and is not comparable to our sequence-only phmmer run.

\section{Embedding geometry and the linear-probe gap}
\label{app:geometry}

\begin{table}[h]
  \caption{Anisotropy and effective dimensionality on the SCOPe-40 corpus, with remote-homology accuracy under three readouts. Participation ratio is over 480 dimensions. Whitening is an unsupervised invertible linear map, and on this task it closes most of the gap between the 3-NN and linear readouts for every model. ProtSent shows no linear-probe deficit on remote homology ($0.701$ against $0.690$), so this table describes the geometry of the space and does not test the cause of the deficit in Section~\ref{sec:results:probes}. The three readout columns come from this analysis rather than from the benchmark sweep, so they differ from the corresponding cells of Table~\ref{app:tab:35m_knn} in the third decimal; the comparison that matters is within a row.}
  \label{app:tab:geometry}
  \centering
  \small
  \begin{tabular}{lccccc}
    \toprule
    Model & Mean random-pair cos & Participation ratio & 3-NN raw & 3-NN whitened & Linear \\
    \midrule
    ESM-2 35M       & 0.848 & 7.9  & 0.584 & 0.680 & 0.690 \\
    ProtSent-V1 35M & 0.294 & 31.0 & 0.658 & 0.677 & 0.691 \\
    ProtSent-V2 35M & 0.152 & 52.5 & 0.667 & 0.701 & 0.701 \\
    \bottomrule
  \end{tabular}
\end{table}

\paragraph{Layer sweep.} Both probes in the main text pool the final layer. Sweeping it on remote homology, ESM-2 35M peaks at layer 6 ($0.670$) and ProtSent-V2 at layer 8 ($0.703$); ProtSent-V2 at its final layer ($0.680$) still exceeds ESM-2 at its best. \emph{These are not the suite's numbers}: the sweep scores a subsample (8{,}000 train, 3{,}000 test) so that five layers could be fitted per model, which sits above the full-split accuracies of Table~\ref{tab:probes} and Section~\ref{sec:results:decontam}. Only the within-sweep comparison across layers is meaningful. The reported advantage is therefore not a layer-choice artifact, but the absolute numbers in the main text are not layer-optimized for anyone.

\paragraph{Macro-F1 on rare classes.} At 150M, ProtSent-V2 minus the frozen base under a linear probe is $-0.026\,[-0.045,-0.007]$ on macro-F1 while accuracy is unresolved ($-0.001\,[-0.011,+0.009]$). The test set has 457 classes with median support 3 and 209 classes with $\leq 2$ examples; restricting to classes with $\geq 3$ test examples shrinks the macro-F1 gap from $-0.026$ to $-0.004$. Roughly $86\%$ of the deficit sits in classes too small to estimate. Against the frozen base, V1 is \emph{worse} than V2 on this metric, although the two differ in more than decontamination.

\section{Seed variability}
\label{app:seeds}

Five probe seeds across 8 tasks and 3 models under a 3-NN probe give a \textbf{median standard deviation of $0.000$} over 24 rows; only thermostability, which subsamples, reaches $0.013$--$0.017$. This is a statement about the probe, which is deterministic on fixed embeddings and a fixed test split. It is \textbf{not} a measure of run-to-run variance, and must not be read as one. One training run exists per model, so training-seed variance is unmeasured throughout this paper. Few-shot variability, where the spread is real, is in Table~\ref{tab:fewshot} and the per-task grid below. The gain there is task-bound: on metal-ion binding at $N{=}1000$ under a linear probe, ESM-2 35M reaches $0.666$ against $0.637$ for V1 and $0.595$ for V2 (Table~\ref{app:tab:fewshot_rest}).

\begin{table}[h]
  \caption{Few-shot remote homology, ESM-2 35M backbone: accuracy, mean (standard deviation) over 5 seeds, fixed estimator, full test split. The other three re-run tasks are in Table~\ref{app:tab:fewshot_rest}.}
  \label{tab:fewshot}
  \centering
  \small
  \begin{tabular}{lcccccc}
    \toprule
    & \multicolumn{3}{c}{3-NN probe} & \multicolumn{3}{c}{Linear probe} \\
    \cmidrule(lr){2-4} \cmidrule(lr){5-7}
    $N$ & ESM-2 & V1 & V2 & ESM-2 & V1 & V2 \\
    \midrule
    50   & 0.061\,(0.010) & 0.055\,(0.008) & 0.045\,(0.009) & 0.121\,(0.003) & 0.159\,(0.004) & 0.145\,(0.005) \\
    100  & 0.115\,(0.007) & 0.135\,(0.008) & 0.125\,(0.024) & 0.222\,(0.006) & 0.282\,(0.007) & 0.258\,(0.009) \\
    250  & 0.148\,(0.002) & 0.223\,(0.011) & 0.200\,(0.010) & 0.310\,(0.007) & 0.394\,(0.012) & 0.368\,(0.013) \\
    1000 & 0.185\,(0.002) & 0.318\,(0.015) & 0.289\,(0.016) & 0.288\,(0.014) & 0.377\,(0.008) & 0.355\,(0.009) \\
    \bottomrule
  \end{tabular}
\end{table}

\begin{table}[h]
  \caption{Few-shot results on the remaining three tasks, ESM-2 35M backbone, mean (SD) over 5 seeds. Stability is the case that motivates reporting absolute scores with seeds: its standard deviation at small $N$ is comparable to the differences between arms, so relative deltas would be uninformative.}
  \label{app:tab:fewshot_rest}
  \centering
  \scriptsize
  \setlength{\tabcolsep}{3pt}
  \begin{tabular}{llcccccc}
    \toprule
    & & \multicolumn{3}{c}{3-NN probe} & \multicolumn{3}{c}{Linear probe} \\
    \cmidrule(lr){3-5} \cmidrule(lr){6-8}
    Task & $N$ & ESM-2 & V1 & V2 & ESM-2 & V1 & V2 \\
    \midrule
    \multirow{4}{*}{Solubility}
      & 50   & 0.518\,(0.005) & 0.497\,(0.003) & 0.488\,(0.002) & 0.536\,(0.005) & 0.524\,(0.009) & 0.519\,(0.006) \\
      & 100  & 0.521\,(0.002) & 0.531\,(0.002) & 0.519\,(0.002) & 0.567\,(0.003) & 0.525\,(0.003) & 0.532\,(0.006) \\
      & 250  & 0.534\,(0.001) & 0.520\,(0.002) & 0.535\,(0.001) & 0.537\,(0.001) & 0.547\,(0.001) & 0.548\,(0.004) \\
      & 1000 & 0.549\,(0.000) & 0.540\,(0.000) & 0.548\,(0.001) & 0.574\,(0.001) & 0.575\,(0.002) & 0.556\,(0.002) \\
    \midrule
    \multirow{4}{*}{Metal ion binding}
      & 50   & 0.524\,(0.002) & 0.541\,(0.003) & 0.515\,(0.007) & 0.554\,(0.004) & 0.526\,(0.003) & 0.539\,(0.006) \\
      & 100  & 0.537\,(0.004) & 0.522\,(0.003) & 0.522\,(0.003) & 0.579\,(0.005) & 0.554\,(0.002) & 0.550\,(0.002) \\
      & 250  & 0.581\,(0.002) & 0.564\,(0.002) & 0.570\,(0.003) & 0.642\,(0.005) & 0.588\,(0.003) & 0.590\,(0.004) \\
      & 1000 & 0.638\,(0.002) & 0.643\,(0.001) & 0.643\,(0.002) & 0.666\,(0.001) & 0.637\,(0.004) & 0.595\,(0.001) \\
    \midrule
    \multirow{4}{*}{Stability}
      & 50   & 0.178\,(0.054) & 0.210\,(0.102) & 0.327\,(0.161) & 0.216\,(0.202) & 0.233\,(0.134) & 0.200\,(0.131) \\
      & 100  & 0.260\,(0.129) & 0.273\,(0.162) & 0.401\,(0.202) & 0.284\,(0.209) & 0.298\,(0.177) & 0.292\,(0.200) \\
      & 250  & 0.226\,(0.145) & 0.282\,(0.115) & 0.343\,(0.130) & 0.373\,(0.145) & 0.399\,(0.169) & 0.439\,(0.106) \\
      & 1000 & 0.228\,(0.074) & 0.228\,(0.071) & 0.315\,(0.079) & 0.406\,(0.075) & 0.471\,(0.092) & 0.430\,(0.105) \\
    \bottomrule
  \end{tabular}
\end{table}

\section{Late interaction: additional results}
\label{app:late}

\paragraph{Dependence on the SCOPe level.} The SCOPe-40 hierarchy supports fold, superfamily and family labels, and \textbf{two of the effects below change sign or lose significance depending on which level is scored}. We report family, to match the rest of the paper, and flag the difference wherever it matters. In particular: at the \emph{superfamily} level, frozen MaxSim gets \emph{worse} with scale for ESM-2 ($-0.029\,[-0.042,-0.016]$ MAP from 35M to 150M) while improving for ProtSent ($+0.036\,[+0.027,+0.045]$), which would license reading scale as an interaction rather than a main effect; at the \emph{family} level both improve ($+0.060$ and $+0.066$) and no such reading is available. Similarly, a 128-dimensional late-interaction projection beats a 640-dimensional one at superfamily ($+0.010\,[+0.007,+0.014]$) and is unresolved at family ($-0.002\,[-0.007,+0.003]$). We therefore make neither claim.

\begin{table}[h]
  \caption{Scoring \emph{identical frozen weights} with MaxSim instead of pooled cosine, SCOPe-40 family retrieval over the 1{,}693 eligible queries. Nothing is retrained and no parameter is added; only the similarity changes. $\Delta$MAP is a paired bootstrap, $5{,}000$ resamples; all four intervals exclude zero.}
  \label{tab:late}
  \centering
  \small
  \begin{tabular}{lccc}
    \toprule
    Model & MAP, pooled cosine & MAP, MaxSim & $\Delta$MAP \\
    \midrule
    ESM-2 35M        & 0.421 & 0.545 & $+0.124\,[+0.113,+0.135]$ \\
    ESM-2 150M       & 0.424 & 0.606 & $+0.182\,[+0.168,+0.196]$ \\
    ProtSent-V2 35M  & 0.646 & 0.684 & $+0.038\,[+0.032,+0.044]$ \\
    ProtSent-V2 150M & 0.705 & 0.750 & $+0.045\,[+0.039,+0.052]$ \\
    \bottomrule
  \end{tabular}
\end{table}

\begin{table}[h]
  \caption{Paired effects on SCOPe-40 average precision at the family level, $5{,}000$ resamples over the 1{,}693 eligible queries. Effects that hold at every level are the only ones the main text uses.}
  \label{app:tab:late_paired}
  \centering
  \small
  \begin{tabular}{lcc}
    \toprule
    Effect ($a - b$) & $\Delta$MAP & 95\% CI \\
    \midrule
    ProtSent-V2 post-training, 35M, frozen MaxSim  & $+0.139$ & $[+0.124,+0.154]$ \\
    ProtSent-V2 post-training, 150M, frozen MaxSim & $+0.144$ & $[+0.130,+0.158]$ \\
    ProtSent-V2 post-training, 35M, pooled cosine  & $+0.225$ & $[+0.210,+0.240]$ \\
    ProtSent-V2 post-training, 150M, pooled cosine & $+0.281$ & $[+0.265,+0.297]$ \\
    \midrule
    MaxSim $-$ cosine, same ESM-2 35M weights      & $+0.124$ & $[+0.113,+0.135]$ \\
    MaxSim $-$ cosine, same ESM-2 150M weights     & $+0.182$ & $[+0.168,+0.196]$ \\
    MaxSim $-$ cosine, same ProtSent-V2 35M weights  & $+0.038$ & $[+0.032,+0.044]$ \\
    MaxSim $-$ cosine, same ProtSent-V2 150M weights & $+0.045$ & $[+0.039,+0.052]$ \\
    \bottomrule
  \end{tabular}
\end{table}

\paragraph{Two-stage retrieval.} Shortlisting by pooled cosine and re-ranking only the shortlist with MaxSim recovers $97.4\%$ of exhaustive MaxSim family MAP at 10 candidates per query, $99.2\%$ at 30 and $99.8\%$ at 100. Only shortlisted candidates need residue-level embeddings, so the stored index stays pooled-size.

\paragraph{Computational cost.} On the 2{,}207-domain gallery, phmmer takes 4.2\,s on 32 CPU threads; pooled cosine over ProtSent-150M takes 9.0\,s on one GPU; exhaustive MaxSim over the same frozen 640-dimensional residues takes 46.8\,s, of which encoding is about a fifth and scoring the rest. Scoring cost scales with dimension, so a trained 128-dimensional projection scores roughly $3\times$ faster, which is the argument for two-stage retrieval rather than exhaustive MaxSim.

\paragraph{Comparison with phmmer at top-1.} Under MaxSim, ProtSent-V2-150M reaches family R@1 $0.774$ against phmmer's filters-off $0.753$. We do not treat this as a result. No paired test is possible, because per-query phmmer hits were not stored; the one paired comparison on record has a CI half-width of about $0.024$, wider than this $0.021$ margin; and phmmer is $5$--$11\times$ faster depending on the MaxSim configuration compared. The defensible statement about that comparison is the coverage one already made in Section~\ref{sec:results:scope}.

\paragraph{Late-interaction training.} Training a dedicated late-interaction projection on top of ProtSent adds $+0.014\,[+0.006,+0.021]$ family MAP at 150M over the frozen model, an interval that excludes zero but a difference no larger than the checkpoint-to-checkpoint spread at that scale (Appendix~\ref{app:structural}). We do not interpret this as an upper bound: that run had not converged. Its SCOPe curve was still rising at the end of training, it saw under a third of one epoch, and a freshly initialized projection head itself costs MAP at step 0, so early training is spent recovering that. The supported claim is that late interaction is beneficial without additional training; whether additional training helps further remains open. None of our late-interaction runs isolates step count: the arms that differ in step count also differ in sampler, pool size, world size, attention backend and compilation, so no claim about training length holds in either direction. Every late-interaction arm is a single seed, and a paired attention-backend A/B moved SCOPe by up to $0.0045$ at $n{=}1$, so effects below about $0.005$ here are not resolved.

\section{Full per-task ablation results}
\label{app:ablation_full}

\textbf{Reporting convention.} Every cell in Table~\ref{tab:ablation_full} is a \emph{relative} change against stock ESM-2 35M, measured on the \textbf{V1 model}, on the suite's \textbf{default split}, from a \textbf{single run}. That convention is weaker than the one used elsewhere in this paper, and we keep it here only because re-running six ablations on the decontaminated corpus was out of reach. These rows support the \emph{direction} of source-specific effects, which task family each source moves, and nothing finer. A second test of the same question, on the decontaminated MNRL corpora, is the V2.5 experiment in Section~\ref{sec:ablations}, subject to the confounds listed in Appendix~\ref{app:training}.

\begin{table}[h]
  \caption{Per-task ablation results (ProtSent-V1, ESM-2 35M, default split, single run, 3-NN probe except the multilabel tasks, which use their built-in linear evaluator). Each cell is the relative change (\%) against stock ESM-2. Bold marks the best configuration per task. The two summary rows cover the 23 single-label tasks and exclude the two multilabel rows.}
  \label{tab:ablation_full}
  \centering
  \scriptsize
  \setlength{\tabcolsep}{3pt}
  \begin{tabular}{l c r r r r r r r}
    \toprule
    Task & Metric & Full & $-$Pfam & $-$HardNeg & $-$AFDB & $-$StringDB & Prop. & $-$DMS \\
    \midrule
    \multicolumn{9}{l}{\textit{Binary}} \\
    PPI & AUC & +5.3 & +5.9 & +4.3 & +5.5 & $-$0.5 & \textbf{+6.6} & +5.9 \\
    Peptide-HLA & AUC & +3.6 & +4.0 & \textbf{+7.0} & +4.0 & +4.9 & +6.2 & +5.0 \\
    Solubility & AUC & +2.7 & +2.5 & +2.6 & \textbf{+2.8} & +1.3 & +2.8 & +2.4 \\
    Neuropeptide & AUC & +0.9 & +0.1 & \textbf{+1.1} & +0.1 & +0.1 & +0.8 & +0.2 \\
    Binary SubCell & AUC & +0.8 & +1.0 & +0.9 & +0.7 & +0.8 & \textbf{+1.4} & +0.8 \\
    Signal Peptide & AUC & +0.7 & +1.0 & +1.2 & $-$0.3 & +0.1 & \textbf{+1.4} & $-$0.1 \\
    Metal Ion & AUC & +0.5 & +1.6 & \textbf{+1.7} & $-$1.8 & +0.7 & +0.9 & +0.7 \\
    Material Prod. & AUC & +0.4 & +0.8 & +0.6 & $-$0.3 & +0.7 & +1.1 & \textbf{+1.2} \\
    \midrule
    \multicolumn{9}{l}{\textit{Multiclass}} \\
    Remote Homology & F1 & +40.5 & +42.5 & +49.5 & +15.3 & +43.3 & \textbf{+54.7} & +40.6 \\
    SubCell Loc. & AUC & +1.3 & +1.8 & \textbf{+2.2} & +1.5 & $-$3.1 & +1.6 & +2.0 \\
    EC Number & F1 & +0.3 & $-$1.3 & \textbf{+7.7} & $-$11.1 & +3.7 & +2.5 & $-$2.1 \\
    Antibiotic Res. & AUC & +0.1 & $-$0.2 & +0.0 & +0.2 & \textbf{+0.7} & +0.6 & $-$0.3 \\
    Temp. Stability & F1 & $-$2.6 & \textbf{$-$1.1} & $-$1.3 & $-$1.2 & $-$9.3 & $-$1.2 & $-$1.9 \\
    \midrule
    \multicolumn{9}{l}{\textit{Multilabel (built-in linear evaluator)}} \\
    CAFA5 & F1 & $-$2.1 & $-$1.7 & $-$1.1 & $-$6.7 & $-$5.6 & \textbf{+0.6} & $-$1.2 \\
    Mol. Function & F1 & $-$4.6 & $-$4.1 & $-$2.4 & $-$7.2 & $-$5.0 & \textbf{+1.0} & $-$3.3 \\
    \midrule
    \multicolumn{9}{l}{\textit{Regression}} \\
    RhlA & Sp. & \textbf{+77.2} & +21.3 & +48.6 & +38.0 & +36.2 & +59.3 & +59.3 \\
    Beta-lactamase & Sp. & +18.5 & +8.1 & +17.8 & +7.2 & \textbf{+19.5} & +9.3 & +8.9 \\
    Fluorescence & Sp. & +15.6 & +20.6 & \textbf{+24.2} & +22.0 & +19.6 & +18.4 & +10.4 \\
    Optimal pH & Sp. & +3.1 & $-$3.7 & \textbf{+5.9} & $-$6.0 & +4.7 & $-$6.6 & +2.2 \\
    Variant (GB1) & Sp. & $-$0.8 & +9.0 & +7.9 & +11.8 & +13.8 & +8.4 & \textbf{+14.7} \\
    Cloning & Sp. & $-$1.0 & $-$1.4 & $-$1.2 & $-$6.9 & \textbf{$-$0.6} & $-$1.4 & $-$0.9 \\
    AAV Fitness & Sp. & $-$1.7 & +0.2 & \textbf{+3.6} & +1.2 & $-$1.5 & $-$0.2 & $-$1.1 \\
    Enzyme Eff. & Sp. & $-$2.6 & $-$1.8 & +0.1 & $-$3.7 & \textbf{+1.3} & $-$0.1 & +0.0 \\
    Stability & Sp. & $-$3.7 & $-$5.4 & \textbf{+1.0} & $-$2.2 & $-$5.8 & $-$2.5 & $-$3.2 \\
    Thermostability & Sp. & $-$5.3 & $-$0.6 & $-$4.4 & $-$3.3 & \textbf{+5.2} & $-$2.1 & $-$11.8 \\
    \midrule
    \textbf{Tasks improved} & & 16/23 & 15/23 & 20/23 & 13/23 & 17/23 & 16/23 & 15/23 \\
    \textbf{Mean $\Delta$\%} & & +6.7 & +4.6 & +7.9 & +3.2 & +5.9 & +7.0 & +5.8 \\
    \bottomrule
  \end{tabular}
\end{table}

The GB1 column requires a comment. Removing the DMS objective \emph{improves} GB1 variant-effect prediction ($-0.8 \rightarrow +14.7$), which is the opposite of what a fitness-supervision story predicts. Taken with the V2.5 result, where re-adding DMS improves stability, AAV, fluorescence, GB1 and beta-lactamase under a linear probe at 35M, the summary the evidence supports is that the DMS objective moves DMS-shaped tasks, but not uniformly and not always in the direction one would guess. We make no general claim linking DMS training to fitness prediction.

% \section{Why no UMAP panels}
% \label{app:umap}

% We do not show UMAP projections of the baseline and post-trained spaces. In our own projections, several SCOPe-40 structural classes appear \emph{more} scattered after post-training than before, the opposite of the quantitative result in Figure~\ref{fig:hierarchy}; Pfam-family panels at 35M are hard to tell apart; and a Pfam-family panel at 150M shows the model separating the classes it was trained to separate, which is not evidence of anything held out. A UMAP layout is also sensitive to its own hyperparameters. Figure~\ref{fig:hierarchy} answers the same question, whether the space organizes itself by biological relatedness, on all $\binom{2207}{2}$ pairs rather than a projection of a color-coded subset, and it is falsifiable in a way a scatter plot is not: a representation that encodes the hierarchy must produce a monotone curve, and the frozen backbone does not.

\section{Limitations (vi) to (x)}
\label{app:limitations}

Limitations (i) to (v), which bear most directly on the central claim, are stated in Section~\ref{sec:discussion}. These five complete the list and continue its numbering.

(vi) \emph{The DMS corpus was never identity-decontaminated, and it does overlap the benchmarks nearest to it.} An identity check at $40\%$ / $80\%$ finds $2.2\%$ of the corpus matching the fluorescence test set, $0.6\%$ the stability set and $0.3\%$ beta-lactamase, at a median identity above $0.9$ among hits; only GB1, whose assays are dropped by name, is clean at $0.0\%$ (Appendix~\ref{app:datasets}). Those three cells on the V2.5 and ESM-C models should therefore be treated as affected by leakage. The primary V2 models do not train on DMS at all.

(vii) \emph{The DMS objective conflates high fitness with wild-type-likeness} (Section~\ref{sec:method:objective}); we keep it auxiliary and report the limitation.

(viii) \emph{Mutation-effect prediction is worse after post-training, in a regime where no embedding method is competitive.} Under frozen MaxSim on ProteinGym, ProtSent-V2 trails stock ESM-2 at both scales, on DMS substitutions ($0.314$ against $0.335$ at 35M, $0.325$ against $0.335$ at 150M, 214 assays) and on indels ($0.524$ against $0.541$; $0.551$ against $0.593$, 66 assays), having led it by $+0.14$ MAP on structural retrieval under the same scoring. No single difference is significant, since every pair of 95\% intervals overlaps, but the direction holds across both variant types and both scales, and a plausible interpretation is that contrastive training on family and structural positives builds exactly the invariance that a point substitution must break in order to be scored. More importantly, the whole comparison sits below the appropriate method: ESM-2 150M scored by standard masked marginals reaches $0.388$ on DMS substitutions, against $0.325$--$0.346$ for \emph{every} 150M MaxSim arm. Embedding similarity, from ProtSent or from the backbone, is not a competitive variant-effect scorer, and we do not propose it as one.

(ix) \emph{Unrun baselines.} ESM-S, S-PLM, Magneton, Foldseek, PLMSearch, DHR, ProTrek and the ProtBERT sentence-transformer of \citet{redl2023optimizing}; no fine-tuning sweep; no joint no-AFDB/no-Pfam ablation. Applying ProtSent to a sequence--structure backbone such as SaProt or ProSST is blocked on data rather than code: both consume residue-level structure tokens we do not have for the Pfam and STRING sequences.

(x) \emph{The late-interaction runs are single-seed}, and none of them isolates step count.

\end{document}